\documentclass[journal]{IEEEtran}
\usepackage{cite}

\ifCLASSINFOpdf
\else
\fi
\usepackage{graphicx}
\usepackage{times}
\usepackage{epsfig}
\usepackage{graphicx}
\usepackage{amsmath}
\usepackage{amssymb}

\usepackage{xcolor}
\usepackage{algorithm}
\usepackage{algorithmic}
\usepackage{pbox}
\usepackage[ruled,vlined,algo2e]{algorithm2e}
\providecommand{\SetAlgoLined}{\SetLine}
\usepackage{tabulary}
\usepackage{multirow}
\usepackage{amsmath}
\begin{document}
%
\title{ZoomCount: A Zooming Mechanism for Crowd Counting in Static Images}
%
%
%
\author{Usman~Sajid,
        Hasan~Sajid,~\IEEEmembership{Member,~IEEE},
        Hongcheng~Wang,~\IEEEmembership{Senior Member,~IEEE,}
        and~Guanghui~Wang,~\IEEEmembership{Senior Member,~IEEE}
\thanks{The work was supported in part by NSF NRI and USDA NIFA under the award no. 2019-67021-28996, and the Nvidia GPU grant.}
\thanks{U. Sajid and G. Wang are with the Department of Electrical Engineering and Computer Science, University of Kansas, Lawrence, KS 66045}
\thanks{H. Sajid is with School of Mechanical and Manufacturing Engineering, National University of Science \& Technology, Pakistan.}
\thanks{H. Wang is with Comcast AI Research, USA.}
\thanks{Manuscript received xxxx; revised xxxx.}}

%
%

\markboth{IEEE Trans. on Circuits and Systems for Video Technology, 2019}%
{Sajid \MakeLowercase{\textit{et al.}}: ZoomCount: A Zooming Mechanism for Crowd Counting in Static Images}
%



\maketitle

\begin{abstract}
This paper proposes a novel approach for crowd counting in low to high density scenarios in static images. Current approaches cannot handle huge crowd diversity well and thus perform poorly in extreme cases, where the crowd density in different regions of an image is either too low or too high, leading to crowd underestimation or overestimation. The proposed solution is based on the observation that detecting and handling such extreme cases in a specialized way leads to better crowd estimation. Additionally, existing methods find it hard to differentiate between the actual crowd and the cluttered background regions, resulting in further count overestimation. To address these issues, we propose a simple yet effective modular approach, where an input image is first subdivided into fixed-size patches and then fed to a four-way classification module labeling each image patch as low, medium, high-dense or no-crowd. This module also provides a count for each label, which is then analyzed via a specifically devised novel decision module to decide whether the image belongs to any of the two extreme cases (very low or very high density) or a normal case. Images, specified as high- or low-density extreme or a normal case, pass through dedicated zooming or normal patch-making blocks respectively before routing to the regressor in the form of fixed-size patches for crowd estimate. Extensive experimental evaluations demonstrate that the proposed approach outperforms the state-of-the-art methods on four benchmarks under most of the evaluation criteria.
\end{abstract}


\ifCLASSOPTIONpeerreview
\begin{center} \bfseries EDICS Category: 3-BBND \end{center}
\fi
%
\IEEEpeerreviewmaketitle

\section{Introduction}
%
%
%
%
\IEEEPARstart{I}{n} recent years, convolutional neural networks have attracted a lot of attention and been successfully applied to various computer vision problems, such as object detection \cite{ma2019mdfn, li2019object, xu2019towards}, face recognition \cite{cen2019dictionary}, depth estimation \cite{he2018learning, he2018spindle}, image classification \cite{zhang2018bpgrad, cen2019boosting}, image-to-image translation \cite{xu2019toward, xu2019adversarially}, and crowd counting \cite{sajid2020plug}. Crowd counting is an integral part of crowd analysis. It plays an important role in event management of huge gatherings like Hajj, sporting, and musical events or political rallies. Automated crowd count can lead to better and effective management of such events and prevent any unwanted incident \cite{helbing2012crowd}.
Crowd counting is an active research problem due to different challenges pertaining to large perspective, huge variance in scale and image resolution, severe occlusions and dense crowd-like cluttered background regions. Also, manual crowd counting subjects to very slow and inaccurate results due to the complex issues as mentioned above.
\begin{figure}[t]
\label{fig:fig1}
	\begin{minipage}[b]{0.490\columnwidth}
		\begin{center}
			\centerline{\includegraphics[width=0.795\columnwidth]{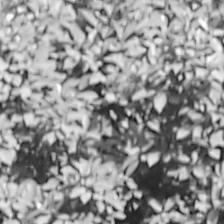}}
			\centerline{\footnotesize{(a) GT=0, Regression=75}}
			\centerline{\footnotesize{Ours=0, Density=72}}
		\end{center}
	\end{minipage}
	\begin{minipage}[b]{0.490\columnwidth}
		\begin{center}
			\centerline{\includegraphics[width=0.795\columnwidth]{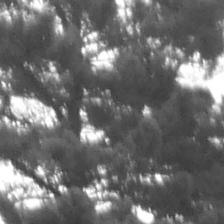}}
			\centerline{\footnotesize{(b) GT=0, Regression=18}}
			\centerline{\footnotesize{Ours=0, Density=19}}
		\end{center}
	\end{minipage}
		\begin{minipage}[l]{0.490\columnwidth}
		\begin{center}
			\centerline{\includegraphics[width=0.995\columnwidth]{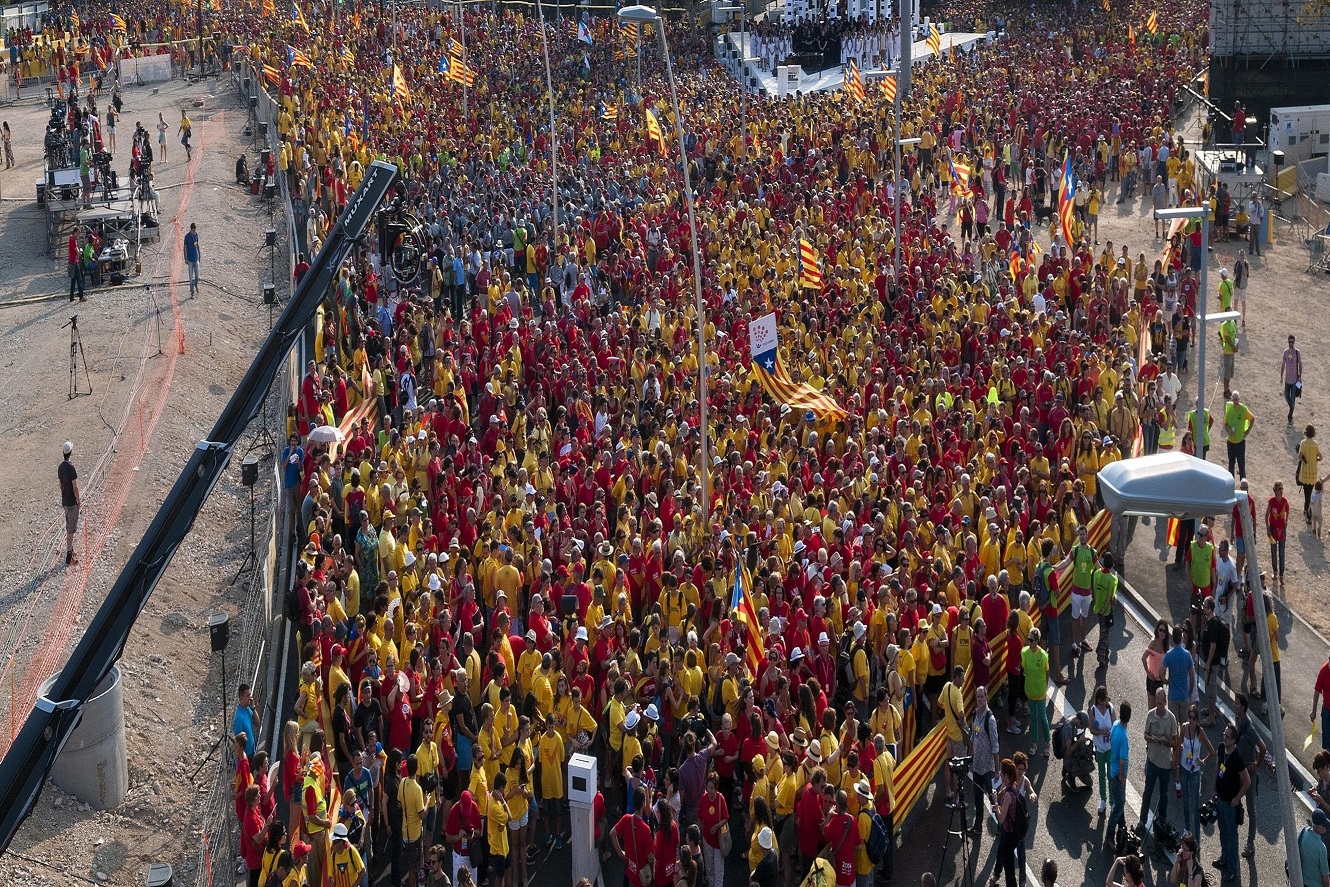}}
			\centerline{\footnotesize{(c) GT=4535, Regression=4109}}
			\centerline{\footnotesize{Z\textsubscript{in}(Ours)=4523, Density=2759}}
		\end{center}
	\end{minipage}
	\begin{minipage}[r]{0.490\columnwidth}
		\begin{center}
			\centerline{\includegraphics[width=0.995\columnwidth]{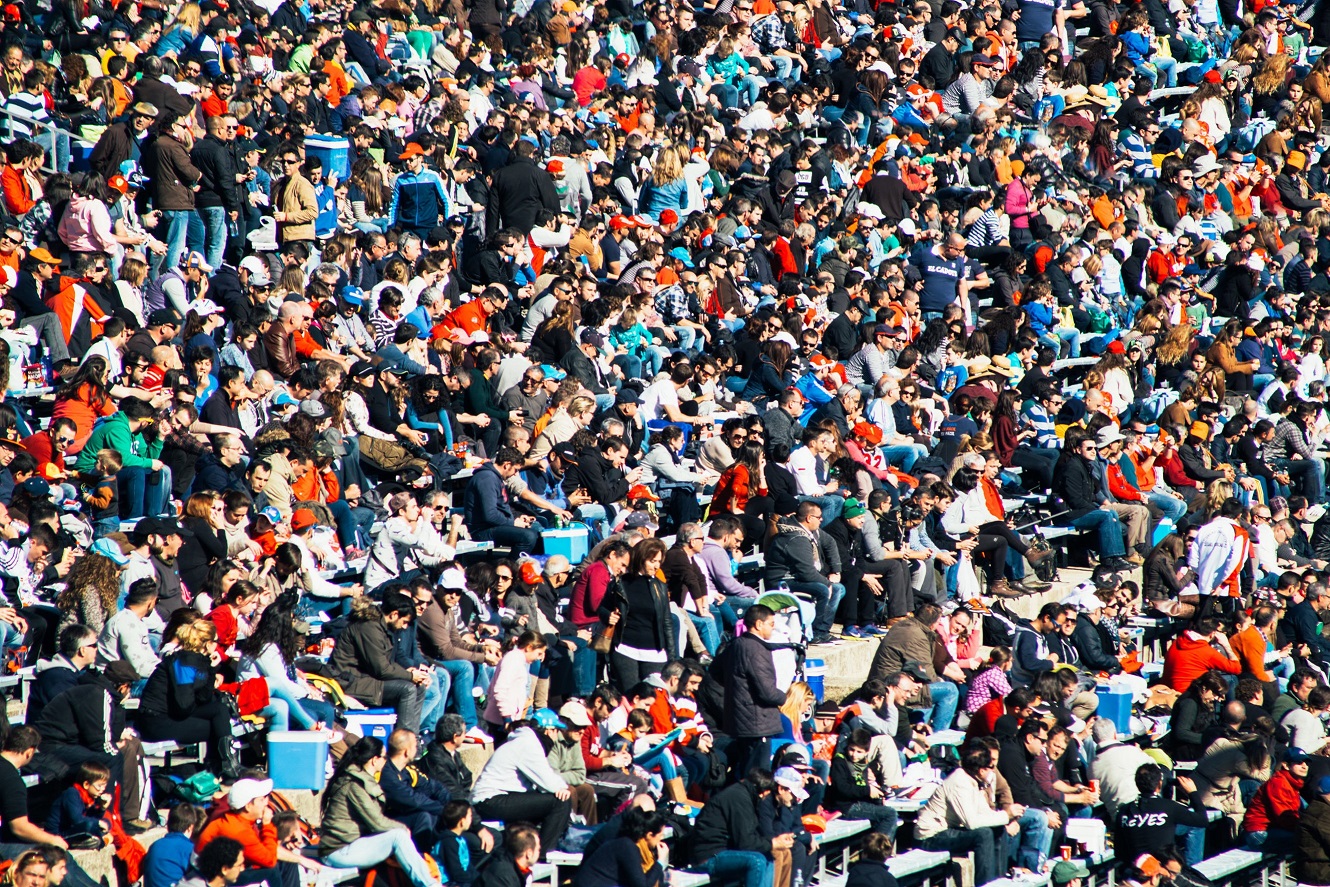}}
			\centerline{\footnotesize{(d) GT=704, Regression=861}}
			\centerline{\footnotesize{Z\textsubscript{out}(Ours)=708, Density=1017}}
		\end{center}
	\end{minipage}			
    \vspace{-0mm}
	\caption{\footnotesize{Direct regression and Density map \cite{ucfonlinedemosite} methods overestimate in case of crowd-like cluttered background patches as in (a) and (b), where there is no crowd at all. Similarly, these methods highly underestimate or overestimate in two extreme cases, where most crowd patches belong to either high or low-crowd count as in (c) and (d) respectively, as compared to the ground truth (GT).
	}}
    \vspace{-3mm}
    
\end{figure}

\begin{figure*}

		\begin{minipage}[b]{2\columnwidth}
		\begin{center}
			\centerline{\includegraphics[width=\columnwidth]{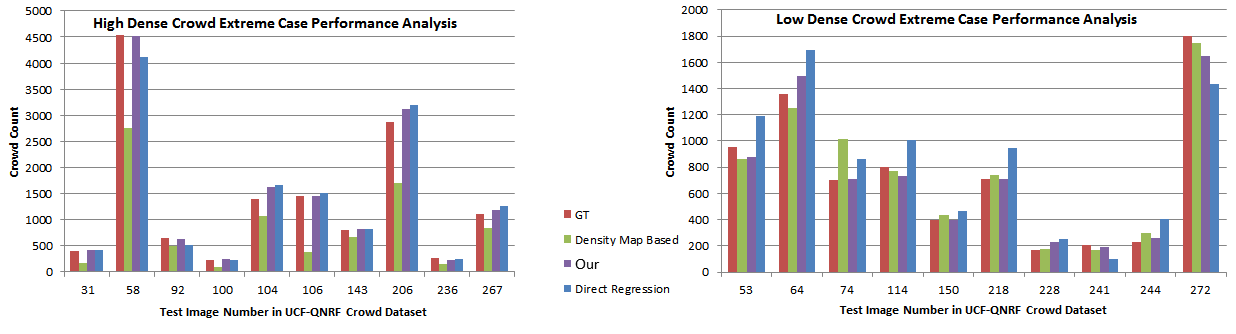}}
		\end{center}
	\end{minipage}
		
    \vspace{-1mm}
	\caption{\footnotesize{Left and right graphs compare ten cases each, belonging to high dense and low dense extreme respectively for Density map \cite{ucfonlinedemosite}, DenseNet \cite{huang2017densely} based direct regression and our method. As shown, other models either highly underestimate or overestimate, whereas the proposed method remains the closest to the ground truth (GT) bar in most cases.
	}}
    \vspace{-3mm}
\label{fig:graphi_ten_cases}    
\end{figure*}

To obtain accurate, fast and automated results, CNN-based approaches have been proposed that achieve superior performance over traditional approaches \cite{chan2009bayesian,chen2012feature,wu2005detection}. CNN-based methods can be broadly classified into three categories; regression-based, detection-based, and density map estimation methods. Regression-based methods \cite{wang2015deep} directly regress the count from the input image. However, these CNN regressors alone cannot handle huge diversity in the crowd images varying from very low to very high. CNN detection-based methods \cite{girshick2015fast,redmon2016you} first detect persons in the image and then sum all detection results to yield the final crowd count estimate. Detection-based methods perform well in low crowd images but could not be generalized well to high-density crowd images as detection fails miserably in such cases due to very few pixels per head or person. Density map estimation methods \cite{cascadedmtl,idrees2018composition,sindagi2017generating} generate density map values, with one value for each image pixel. The final estimate is then calculated by summing all density map values. These methods do not rely on localizing crowd but rather on estimating crowd density in each region of the crowd image. Density map estimation methods outperform other approaches and recent state-of-the-art methods mostly belong to this category. However, density per pixel estimation remains a huge challenge as indicated in \cite{ranjan2018iterative} due to large variations in the crowd density across different images. This naturally leads to a question: \textit{In which scenarios these methods may fail and why?}

One key issue with regression and density map methods is that they only rely on direct count estimate and density map estimation per pixel for the input image respectively, thus, they may get subjected to large crowd count for cluttered background image patches. As shown in Fig. 1, models \cite{idrees2018composition} based on these methods consider this $224\times224$ image patch as a crowd patch and make false estimates, making the system unreliable as similar patterns are bound to occur in many practical scenarios.

\begin{figure}[t]

	\begin{minipage}[b]{\columnwidth}
		\begin{center}
			\centerline{\includegraphics[width=0.995\columnwidth]{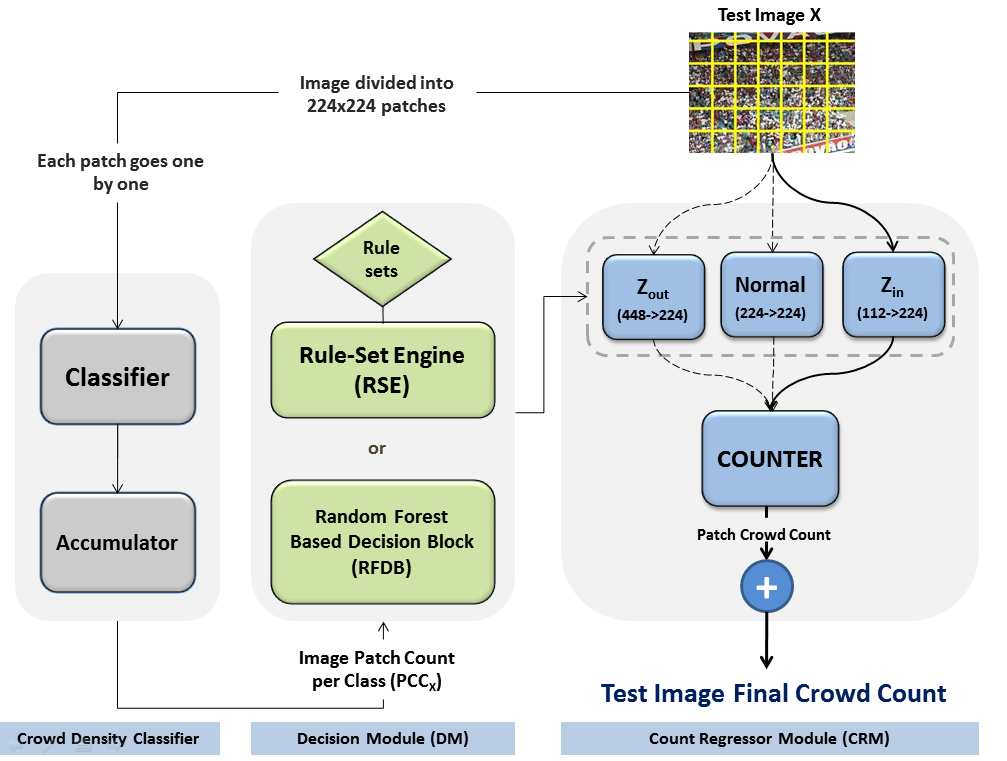}}
		\end{center}
	\end{minipage}		
    \vspace{-7mm}
	\caption{ZoomCount architecture. The test image $X$, divided into $224\times224$ patches, first passes through the Crowd Density Classifier (CDC) module, which discards  no-crowd patches as classified by the robust 4-way DenseNet classifier. The accumulator stores patch count per class $(PCC_X)$ of this image. Decision Module (DM), based on CDC module output and using either autonomous RFDB or heuristic-based RSE module, decides whether this image should be divided into all normal ($Normal$) patches or make all either zoom-in ($Z_{in}$) or zoom-out ($Z_{out}$) based patches before proceeding to the patch-based regressor $(COUNTER)$ for each patch crowd count. Image final crowd estimate is then obtained by summing all patches count. }
    \vspace{-3mm}
    \label{fig:fig4}
\end{figure}

In addition, we observe that both types of methods perform well for images which contain most crowd patches with neither low nor high crowd density. Problem arises when images have most crowd patches with either high or low-density crowd. Due to the limitation in handling such practical diversity in crowd density, these methods may either highly underestimate or overestimate the crowd count in these two extreme cases, as shown in Fig. 1. To further explain this phenomenon, we analyze ten such cases for both extremes separately from very recent UCF-QNRF dataset \cite{idrees2018composition} on the state-of-the-art density map method \cite{ucfonlinedemosite,idrees2018composition} and direct regression-based method as shown in Fig. \ref{fig:graphi_ten_cases}. It can be observed that, in both extreme cases, the crowd estimates are either highly overestimated or underestimated due to the limitations as discussed above.

To solve these fundamental problems, we propose a modular approach as shown in Fig.~\ref{fig:fig4}. It comprises of a Crowd Density Classifier (CDC), a novel Decision Module (DM), and a Count Regressor Module (CRM). The input image is first sub-divided into fixed-size patches ($224\times224$) and fed to the CDC module that contains a deep CNN classifier to perform a four-way classification (low, medium, high-density, and no-crowd) on each patch. The classification module eliminates any crowd-like background patches (no-crowd) from the test image and feeds the information about the number of patches belonging to each of the no-crowd, low, medium and high-density classes to the Decision Module (DM) using an accumulator. DM uses either the machine learning based RFDB module or heuristic-based Rule-Set Engine (RSE) module to determine if the image belongs to a case of low, normal, or high density. Based on the DM decision, this image is then divided into fixed-size patches using one of three independent image patch-making modules $(Z_{in},Normal,Z_{out})$. The image that belongs to low-density extreme case is divided into patches using the zoom-out ($Z_{out}$) patch-maker; the image that belongs to high-density extreme case is divided via zoom-in ($Z_{in}$) patch-making block; and the normal case image is split into patches using normal ($Normal$) patch-maker. These patches are then routed one by one to the patch-wise count regressor $(COUNTER)$  for crowd estimate and the image total crowd count is obtained by summing all patches count.

The $Z_{in}$ block divides each input patch into four $112\times112$ patches, and then up-scales each patch by $2\times$ before routing each patch to the count regressor. Intuitively, this module is further zooming-in into the image and looking in-detail all patches by using $1/2$ input patch size instead of the original $224\times224$ patches. Similarly, zoom-out patch-maker divides the input image into $448\times448$ patches, and down-scales each patch by $2\times$ as it is dealing with the image containing low-density crowd patches mostly. The normal case image directly employs the original $224\times224$ patch size with no up-scaling or down-scaling.
The main contributions of this work include:

\begin{itemize}\setlength\itemsep{-0.3em}
  \item The paper reveals and analyzes the fact that extremely high and low dense crowd images greatly influence the performance of the state-of-the-art regression and density map based methods for crowd counting.
\\  
  \item A novel strategy is proposed to address the problem of counting in highly varying crowd density images by first classifying the images into either one of the extreme cases (of very low or very high density) or a normal case, and then feeding them to specifically designed patch-makers and crowd regressor for counting.
\\  
  \item A novel rule-set engine is developed to determine whether the image belongs to an extreme case. For images of extremely high density, a zoom-in strategy is developed to look into more details of the image; while for images of low-density extreme, a zoom-out based regression is employed to avoid overestimate.
\\
  \item We created four new datasets, each from the corresponding crowd counting benchmark, for the training and testing of different machine learning algorithms to classify an image as normal, high or low-dense extreme case using its patches classification count. These manually verified datasets will facilitate the researchers in analyzing complex crowd diversity, which is at the core of the crowd analysis.

\end{itemize}

The proposed ZoomCount scheme is thoroughly evaluated on four benchmarks: UCF-QNRF \cite{idrees2018composition}, ShanghaiTech \cite{zhang2016single}, WorldExpo'10 \cite{zhang2015cross}, and AHU-Crowd \cite{hu2016dense}. The experimental results demonstrate the effectiveness and generality of the proposed strategy and rule-sets, which are never realized for crowd counting. The overall performance of the proposed model outperforms the state-of-the-art approaches on most of the evaluation criteria. The proposed models and source code, as well as the created datasets, will be available on the author's website.


\section{Related Work}
\label{relatedWork}
Crowd counting is an active research area in computer vision with different challenges related to large perspective, occlusion, cluttered background regions and high variance in crowd density across different images. Earlier work \cite{wang2011automatic,wu2005detection} focused on the head or full-body detection for counting using handcrafted features for detectors learning. These methods failed in case of high dense images, where it is hard to find such handcrafted features. The approaches were shifted towards regression based counting \cite{chan2009bayesian,chen2012feature,ryan2009crowd}, where a mapping function was learned to directly regress count from local patches of an image. These methods improved the counting process, however, they could not handle huge crowd diversity and also lack awareness about crowd density across all parts of the image.

Recently, CNN-based approaches have been widely used \cite{idrees2018composition,li2018csrnet,wang2015deep,liu2018decidenet}. They are broadly categorized into three classes; Counting by detection, counting by direct regression, and counting using density map estimation. CNN-based object detectors \cite{girshick2015fast,redmon2016you} detect each person in the image, and the final count is then calculated by summing all detections. These methods \cite{shami2018people,li2019headnet} deteriorate in high density and severe occlusion cases, where each head only occupies a few pixels. Counting by direct regression methods \cite{wang2015deep} directly regress count by learning feature maps from the input image patch. Wang {\it et al.} \cite{wang2015deep} proposed an end-to-end AlexNet \cite{krizhevsky2012imagenet} based regressor for crowd count. These methods alone cannot handle huge diversity in different crowd images.

Density map estimation methods \cite{idrees2018composition,li2018csrnet,cao2018scale,cascadedmtl,liu2018decidenet} learn to map crowd density per pixel of an image without localizing the counts. The final estimate is calculated by summing all density estimations. Zhang {\it et al.} \cite{zhang2016single} proposed a three-column CNN architecture (MCNN) to handle crowd diversity across images. Each column is designed to handle different scales using different receptive field sizes. Sindagi {\it et al.} \cite{sindagi2017generating} extended the idea of MCNN to incorporate contextual information for high-quality density maps generation. Recently, Sam. {\it et al.} \cite{sam2017switching} proposed SwitchCNN which routes each input patch to one of three independent CNN regressors  using a switch CNN classifier. Based on the classification and regression idea, Sindagi {\it et al.} \cite{cascadedmtl} designed a Cascaded-MTL that estimates count for the whole image by using cascaded 10-way classification prior and final density map estimation. 

Crowd counting models, based on whole image estimation and training from the scratch, are subjected to over-fitting due to limited dataset availability (only a few hundred training images). Thus, patch-based models are widely used nowadays. The final sum is computed by adding up all patch count estimates. Liu {\it et al.} \cite{liu2018decidenet} proposed a hybrid approach by incorporating both regression and detection blocks using an attention-guided mechanism to handle low and high-density cases simultaneously. Li {\it et al.} \cite{li2018csrnet} designed a CSRNet to get multi-scale contextual information by incorporating dilation-based convolutional layers. Idress {\it et al.} \cite{idrees2018composition} proposed a composition loss based model for simultaneous crowd counting and localization. Existing methods perform worse in extreme cases where most crowd patches belong to either high density or low density. Moreover, these methods lack the ability to fully discard any cluttered background regions in the image, thus resulting in overestimate.
\section{Proposed Approach}
\label{proposedApproach}
The proposed framework is shown in Fig.~\ref{fig:fig4}, which is composed of three modules namely Crowd Density Classifier (CDC), Decision Module (DM) and Count Regressor Module (CRM). The input image is first sub-divided into $224\times224$ size patches and each patch then passes through the CDC module for 4-way classification (low, medium, high-density or no-crowd). The accumulator gathers and feeds patch count per class information to the Decision Module. Based on accumulator information and utilizing either Random Forest based Decision Block (RFDB) or heuristic-based Rule-Set Engine (RSE), DM routes this image to one of three specialized patch-making blocks $(Z_{in},Normal,Z_{out})$ of CRM where the input image is divided into corresponding patches, followed by the crowd estimate for each patch via crowd regressor $(COUNTER)$. Finally, the image crowd count is calculated by summing all patches count. Below we will discuss the details of each module, as well as the rules defined for the two possible extremes.
\subsection{Crowd Density Classifier (CDC) Module}
\label{IPCM}

The CDC module is composed of a deep CNN 4-way classifier that specializes in making a distinction between no-crowd (NC), low-density (LC), medium-density crowd (MC), and high-density crowd (HC) for each input patch. Let $X$ be a test image sub-divided into $N$ patches $[x_1,x_2,...x_N]$, each with a size of $224\times224$. The accumulator gathers each patch classification result for the input image $X$ as follows:

\begin{figure}[t]
	\begin{minipage}[b]{0.242\columnwidth}
		\begin{center}
			\centerline{\includegraphics[width=0.995\columnwidth]{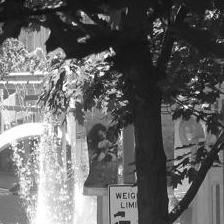}}
			\centerline{\footnotesize{no-crowd(NC)}}
		\end{center}
	\end{minipage}
	\begin{minipage}[b]{0.242\columnwidth}
		\begin{center}
			\centerline{\includegraphics[width=0.995\columnwidth]{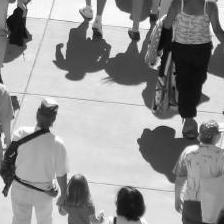}}
			\centerline{\footnotesize{low-crowd(LC)}}
		\end{center}
	\end{minipage}
		\begin{minipage}[b]{0.242\columnwidth}
		\begin{center}
			\centerline{\includegraphics[width=0.995\columnwidth]{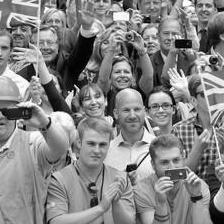}}
			\centerline{\footnotesize{medium-crowd(MC)}}
		\end{center}
	\end{minipage}
	\begin{minipage}[b]{0.242\columnwidth}
		\begin{center}
			\centerline{\includegraphics[width=0.995\columnwidth]{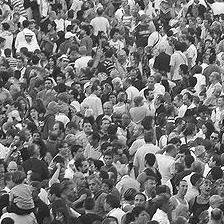}}
			\centerline{\footnotesize{high-crowd(HC)}}
		\end{center}
	\end{minipage}
			
    \vspace{-5mm}
	\caption{\footnotesize{Actual patches being used for the CDC classifier training. They belong to one of the four class labels (NC, LC, MC, HC) based on the definition.
	}}
    \vspace{-3mm}
    \label{fig:labelsexamples}
\end{figure}

\begin{equation}
\label{eq1}
P_{y}\ +=\ 1,\ \ if\ class(x_i)\ =\ y
\end{equation}
for $i=1,2,....N$ and $y$ belongs to either NC, LC, MC or HC class label.
In the end, the accumulator passes the patch count per class ($PCC_{X}$) of this image to the decision module (DM) as:  
\begin{equation}
\label{eq5}
PCC_X\ =\ \{P_{NC},P_{LC},P_{MC},P_{HC}\}
\end{equation}
where $P_{NC},P_{LC},P_{MC}$ and $P_{HC}$ denote the total number of patches being classified as NC, LC, MC and HC respectively of the image $X$. Patches being classified as NC are discarded, and thus remaining $\{N-P_{NC}\}$ crowd patches are going to be used for final crowd estimate. As a result, the crowd-like cluttered background regions (such as the tree leaves shown in Fig. 1), which may result in overestimation otherwise, will be eliminated.

\textbf{Definitions of NC, LC, MC and HC class labels.} During experiments for each crowd counting benchmark dataset, we randomly extract patches from its training images for the CDC classifier training and assign a ground truth class label (NC,LC,MC,HC) to each extracted patch. Since these datasets also contain the localization of people, so we generate the ground truth class label for each patch using this information and the maximum people count possible in any image patch of the corresponding dataset. LC class label is assigned to a patch if the ground truth people count for that patch is less than or equal to 5\% of the maximum possible count but greater than zero as zero crowd means NC class patch. Similarly, patches with ground truth people count between 5\% to 20\% of the maximum possible count are assigned the MC class label, while patches containing more than 20\% of the maximum people count are labeled as HC category patches. In the end, a total of $90,000$ patches, with an equal amount per class label, are generated for the CDC classifier training in each benchmark setting. Example patches for each class label are shown in Fig. \ref{fig:labelsexamples}.

\begin{figure}[t]

	\begin{minipage}[b]{\columnwidth}
		\begin{center}
			\centerline{\includegraphics[width=0.995\columnwidth]{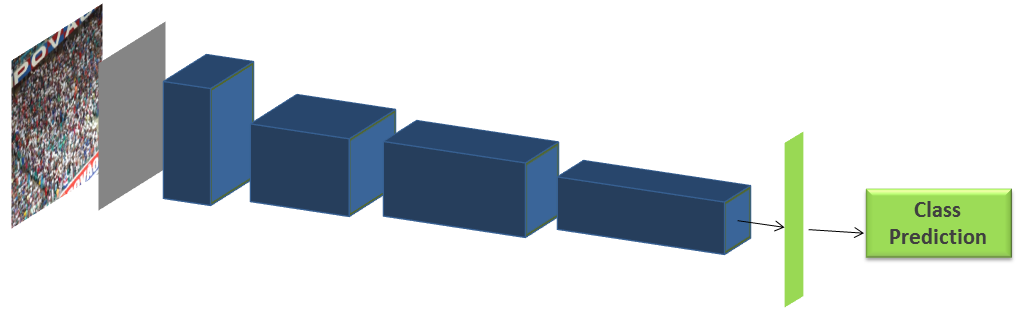}}
		\end{center}
	\end{minipage}		
    \vspace{-7mm}
	\caption{Densenet-201 architecture used for 4-way crowd density classification. Blue blocks represent Dense Blocks 1, 2, 3 and 4 from left to right, followed by a fully connected layer and final softmax 4-way classification. $224\times224$ size input patch is expected.}
    \vspace{-0mm}
    \label{fig:densenet}
\end{figure}

\begin{table*}[t]\small
	\caption{\footnotesize Description of two Rule-Sets: the lower density extreme (Rules 1-4) and the higher density extreme (Rules 5-8). Third column indicates images that are affected the most (in terms of resolution) by that rule. Some rules have much higher tendency to be applied on the Lower Resolution (LR) or Higher Resolution (HR) images, whereas some rules have impact on all types of images (indicated by 'Mix'). }
	
	\begin{center}
	\begin{tabulary}{\linewidth}{|>{\centering\arraybackslash}m{0.09\linewidth}|>{\centering\arraybackslash}m{0.04\linewidth}|>{\centering\arraybackslash}m{0.11\linewidth}|L|}
    \hline
\textbf{Extreme Case Type} & \textbf{Rule} & \textbf{Most Affected Images} & \textbf{Description}\\ \hline

\multirow{4}{*}{Low density}&1 & LR &  Image contains $LC$ and $NC$ patches only.  \\ \cline{2-4}
&2  & Mix & Image should have LC patches and no HC patch.    \\  \cline{2-4}
&3  & HR & Image has more than $50\%$ patches being classified as $LC$ category. \\  \cline{2-4}
&4  & Mix &  At most $5\%$ patches belong to $HC$ category with at least one patch from $NC$ category.     \\ \hline \hline
\multirow{6}{*}{High density}&5 & Mix &  Image with all patches belonging to $HC$ category only.  \\   \cline{2-4}
&6  & Mix &  All patches are $MC$ category only.     \\ \cline{2-4}
&7 & LR &  More than $50\%$ patches of the image are from $HC$ category.  \\ \cline{2-4}
\center{}&\center{8} &\center{Mix} &Image should have $NC$ patches and at least $33\%$ or more from both $P_{HC}$ and $P_{MC}$ category each. Intuitively, first condition of R8 emphasizes the fact that more no-crowd patches shift image towards high dense case, if supported by other given conditions.    \\ \hline
	\end{tabulary}
	\end{center}
	\label{table:rulesTable}
    \vspace{0mm}
\end{table*}

\textbf{Classifier Details.} We use DenseNet-201 \cite{huang2017densely} as our 4-way classifier, as shown in Fig. \ref{fig:densenet}. It has four dense blocks with transition layers (convolution and pooling) in between them to adjust feature maps size accordingly. The DenseNet-201 has consecutive $1\times1$ and $3\times3$ convolutional layers  in each dense block in $\{6,12,48,32\}$ sets respectively. At the end of the last dense block, a classification layer is composed of $7\times7$ global average pooling, followed by $1000-D$ fully connected layer and the final 4-way softmax classification with cross-entropy loss.

\subsection{Decision Module (DM)}
The decision module, based on the CDC module output $PCC_X$, decides if the test image should be treated as a normal image or a low or a high-density extreme case image. DM makes this decision by utilizing one of the two separate and independent decision-making blocks, namely Rule-Set Engine (RSE) and Random Forest based Decision Block (RFDB). RSE is a novel heuristic-based approach which employs the rule-sets to detect if the test image is either an extreme or a normal case, while RFDB is an automated decision-making block based on Random Forest algorithm that learns to map the test image features ($P_{NC},P_{LC},P_{MC},P_{HC}$) to the respective class label ($Z_{in}$, $Normal$, $Z_{out}$). We also create new RFDB training datasets, each from corresponding crowd counting benchmark, for the training of RFDB module as explained in Sec. \ref{rfdataset}. 

\subsubsection{Rule-Set Engine (RSE)}

\begin{algorithm}[t]
    \caption{Rule-Set Engine Algorithm}
    \label{alg:algorithm1}
\SetAlgoLined
\KwIn{$PCC_X$(Patch Count per Class for Test Image $X$)= \{P\textsubscript{\textbf{NC}}, P\textsubscript{\textbf{LC}}, P\textsubscript{\textbf{MC}}, P\textsubscript{\textbf{HC}}\}}
    
\KwOut{Normal or Z\textsubscript{\textbf{in}} or Z\textsubscript{\textbf{out}}}
    
Let $P_{all}=P_{NC}+P_{LC}+P_{MC}+P_{HC}$

\If($Output=Z_{out}$){input patch count satisfies any of following rules}
{
\textbf{Rule 1}: $if\ P_{HC}+P_{MC}==0$

\textbf{Rule 2}: $if\ P_{HC}==0\ and\ P_{LC}>0$
    
\textbf{Rule 3}: $if\ P_{LC}>(P_{all}*0.50)$
    
\textbf{Rule 4}: $if\ P_{NC}>0\ and\ P_{HC}<=(P_{all}*0.05)$

}

\ElseIf ($Output=Z_{in}$){input patch count satisfies any of following rules}
{
    
\textbf{Rule 5}: $if\ P_{LC}+P_{MC}==0$
    
\textbf{Rule 6}: $if\ P_{LC}+P_{HC}==0$

\textbf{Rule 7}: $if\ P_{HC}>(P_{all}*0.50)$

\textbf{Rule 8}: $if\ P_{NC}>0\ and\ P_{MC}>=(P_{all}*0.33)\ and\ \\P_{HC}>=(P_{all}*0.33)$

}
\textbf{else} $Output=Normal$
    
\end{algorithm}

The accumulated patch count per class $(PCC_X)$ from CDC module is tested against two different rule-sets to determine if an input image is a case of low or a high density extreme or a normal one so that it can be divided into patches using the most suitable patch-making block $(Z_{in},Normal,Z_{out})$. The overall goal of RSE is to encourage an image with more number of high-density patches to pass through zoom-in patch-making block $(Z_{in})$, whereas the image with more number of low-density patches goes through a zoom-out patch-making block $(Z_{out})$. If the image does not belong to any of the two extreme cases, it will be treated as a normal case that uses the normal patch-maker $(Normal)$.

\textbf{Rules.} The RSE module consists of two generalized rule-sets, aiming to detect the images belonging to any of the two extreme cases: the low-density extreme (Rules 1-4) and the high-density extreme (Rules 5-8). As illustrated in Algorithm \ref{alg:algorithm1}, if no rule applies to the test image $X$, it will use $Normal$ patch-maker, whereas the image satisfying any rule from $(1-4)$ or $(5-8)$ will generate its patches using $Z_{out}$ or $Z_{in}$ patch-making blocks respectively. Each rule is explained in detail in Table \ref{table:rulesTable}. This table also shows the most affected images by a specific rule in terms of resolution. For example, Rule 7 is highly applicable on relatively lower resolution (LR) images, whereas Rule 2 can affect images of any resolution equally. It is important to note that these rule-sets are used consistently and evaluated across all four publicly available datasets in the experiments, thus demonstrating the generality and efficacy of such rule-sets. In addition, the current rule sets are extendable by adding more rules to refine the classification/decision process. Please note that all parameters in Table \ref{table:rulesTable} are chosen empirically.


\subsubsection{Random Forest based Decision Block (RFDB)}

The scalable rule-sets based decision process yields promising results as demonstrated throughout the experiments in Sec. \ref{experiments}. Nevertheless, there are many heuristics to handle and it requires manual input and special attention while inducting new rules. To address this issue, we propose an automated machine learning based approach that learns the decision process by mapping the four features $(P_{NC({\%})},P_{LC({\%})},P_{MC({\%})},P_{HC({\%})})$ to respective class label ($Z_{in},\ Normal\ or\ Z_{out}$) for each image, where the features denote percentages instead of total image patches belonging to NC, LC, MC and HC classes respectively and labels represent zoom-in, normal and zoom-out patch-making blocks required to generate the patches from the particular input image before proceeding to the count regressor. We employ percentages for features because of the huge variance in resolution across different images in a dataset, which directly influences the features and hence training quality. In addition, since there is no such dataset available for the crowd counting problem to-date that can help in learning this  mapping, thus we generate a new RFDB training dataset from each corresponding benchmark as explained in detail in next subsection. To automate the decision block process, we explored different machine learning classification models and found the random forest-based model to be the most effective as demonstrated in the experiments in Sec. \ref{experiments}. Thus, we choose the random forest algorithm and hence named this module as Random Forest based Decision Block.

Random Forest (RF), being a bootstrap aggregation or bagging based ensemble method, can be used both for classification and regression. We employ the RF algorithm to classify the four features $(P_{NC({\%})},P_{LC({\%})},P_{MC({\%})},P_{HC({\%})})$ to a class label of ($Z_{in}$, Normal or $Z_{out}$) by building, training and tuning a large collection of de-correlated binary decision trees. Each tree then casts a vote for class prediction for the test sample. Finally, the class label with a majority vote is assigned to that test sample i.e., the input image.

Each RF decision tree $t_k$ is built using a bootstrap sample $BS(t_k)$ which is generated from the training data. Such bootstrap sample is given as:
\begin{equation}
\label{eq_rf1}
BS (t_k) = \begin{bmatrix}
    NC_1 & LC_1 & MC_1 & HC_1  & C_1 \\
    NC_2 & LC_2 & MC_2 & HC_2  & C_2 \\
    NC_3 & LC_3 & MC_3 & HC_3  & C_3 \\
    \vdots & \vdots & \vdots & \vdots & \vdots \\
    NC_M & LC_M & MC_M & HC_M  & C_M \\
\end{bmatrix}
\end{equation}
for $K=0,1,2,...N-1$, where N denotes the total number of RF trees. Each row represents one training sample for the tree $t_k$ with the class label as the last entry. We use $N=100$, which is set empirically as no significant improvement has been observed in performance beyond this number. The trees are grown using the classification and regression (CART) algorithm, where the nodes get split until all leaves become unmixed or contain less than $m_{min}$ samples \cite{scikitrfonline}. We use $m_{min}=2$ throughout our experiments, thus splitting nodes until they contain either only one sample or become pure. To quantify the quality of a tree node split, {\it Gini Impurity} has been used as:
\begin{equation}
\label{eq_rf2}
{Gini\ Impurity}_n = \sum_{i=1}^{L=3} -F_{i}(1-F_{i})
\end{equation}
where $L$ denotes the total unique class labels and $F_i$ denotes the frequency of class label $i$ at node $n$. During testing, each RF tree gives its class prediction for test image $X$. Final class label is obtained by the majority vote criterion \cite{hastie2009elements} as follows:
\begin{equation}
\label{eq_rf3}
C_{RF} (X) = majority\ vote {\{C_k(X)\}}_{1}^N
\end{equation}
where $C_k(X)$ represents the class prediction by the $k^{th}$ RF tree.

\begin{figure}[t]
	\begin{minipage}[b]{\columnwidth}
		\begin{center}
			\centerline{\includegraphics[width=0.995\columnwidth]{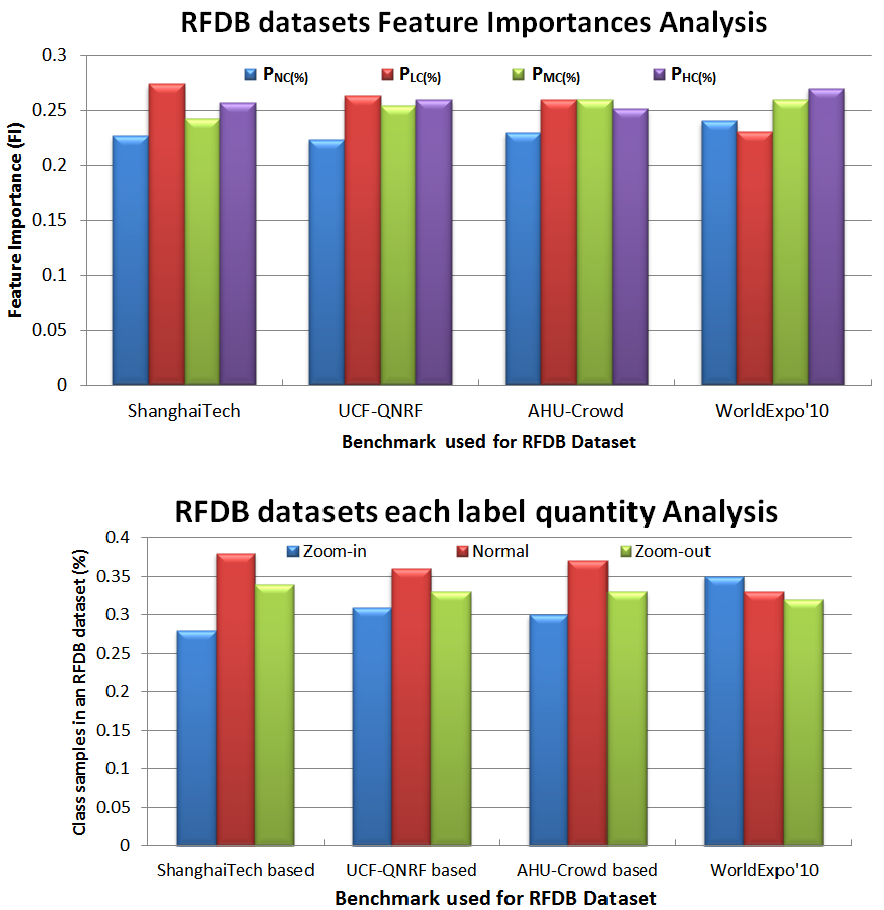}}
		\end{center}
	\end{minipage}		
    \vspace{-7mm}
	\caption{Upper graph shows the Feature Importance (FI) analysis of features from new RFDB datasets created using the corresponding benchmarks (ShanghaiTech, UCF-QNRF, AHU-Crowd, and WorldExpo'10). As shown by the FI value, each of the four features plays an important and equal role in maintaining its respective RFDB dataset quality. Lower graph depicts the total samples per class $(\%)$ in the new RFDB datasets, each created from the corresponding benchmark dataset as indicated by the horizontal axis.}
    \vspace{-3mm}
    \label{fig:fi}
\end{figure}

Feature Importance (FI) depicts the role of each feature in determining the node split and eventually the quality of the RF decision trees building. Features with much lesser FI value can be easily discarded as they do not play any significant role in decreasing the node impurity. As shown in the left graph in Fig. \ref{fig:fi}, all four features have approximately the same FI values in each RFDB dataset. Thus, we keep and use all four available features $(P_{NC({\%})},P_{LC({\%})},P_{MC({\%})},P_{HC({\%})})$ in all four newly generated RFDB datasets.

\subsubsection{Dataset generation for RFDB}
\label{rfdataset}
The RFDB module learns to map the image extracted features $(P_{NC({\%})},P_{LC({\%})},P_{MC({\%})},P_{HC({\%})})$ to the respective class label ($Z_{in},\ Normal\ or\ Z_{out}$) using training dataset with the corresponding mapping. No such dataset has been created to-date. Thus, for each benchmark (ShanghaiTech \cite{zhang2016single}, UCF-QNRF \cite{idrees2018composition}, AHU \cite{hu2016dense}), we created a new respective RFDB dataset which contains this mapping.

To create the new RFDB dataset, each training image's required features $(P_{NC({\%})},P_{LC({\%})},P_{MC({\%})},P_{HC({\%})})$ are extracted using ground truth crowd localization information and definitions of class labels ($NC,LC,MC,HC$) as stated in \ref{IPCM}, followed by manual verification and ground truth (GT) class label assignment. To ensure the quality of the generated dataset, each sample entry was then double checked for any inconsistency, duplicates, missing and erroneous cases. For the extracted features, we use percentages instead of the actual number of patches $(P_{NC},P_{LC},P_{MC},P_{HC})$ belonging to each category because of the huge resolution difference across the images within each benchmark dataset.

\textbf{Statistics.} For each of the four crowd counting benchmarks, we create the corresponding RFDB dataset using its corresponding training images. For instance, in the case of ShanghaiTech dataset (300 training images), we generate the new 300 samples RFDB dataset with each entry being created using one of the respective training image, followed by manual verification that also includes removal or modification of inconsistent entries. In total, 220, 2830 and 812 samples are finalized for the three RFDB datasets based on ShanghaiTech \cite{zhang2016single}, WorldExpo'10 \cite{zhang2015cross} and UCF-QNRF \cite{idrees2018composition} benchmarks, respectively. For AHU \cite{hu2016dense} based RFDB dataset, 90 out of 96 available entries are kept on average with 5-fold cross-validation. The lower graph in Fig. \ref{fig:fi} shows the percentage of each class label in all four newly created RFDB datasets.


\subsection{Count Regressor Module (CRM)}

The CRM module comprises of three independent patch-making blocks and a deep CNN count regressor $(COUNTER)$. The decision module routes the test image to one of these patch-makers for dividing it into $224\times224$ patches after required up-scaling or down-scaling, followed by the crowd count for each image patch via the count regressor $(COUNTER)$. The regressor employs DenseNet-201 \cite{huang2017densely} inspired architecture with a single neuron after the fully connected layer to directly regress the crowd count. Mean squared error (MSE), as defined below, has been employed as the loss function for the count regressor $c$ : 
\begin{equation}
\label{eq2}
L_c = \frac{1}{N} \sum_{i=1}^{N} (F(X_i,\Theta)-Y_i)^{2}
\end{equation}
where $N$ is the number of training patches per batch, $Y_i$ is the ground truth crowd count for the input patch $X_i$, and $F$ is the function that maps the input patch $X_i$ to the crowd count with learnable parameters $\Theta$.

\textbf{Zoom-in based Patch Maker} $(Z_{in})$: Ideally, the decision module $(DM)$ routes the image, with most crowd patches being classified as high-density crowd, to this patch-maker. The image, using this patch-maker, is further sub-divided into equal $112\times112$ patches, and then up-scaled by $2 \times$ before proceeding to the count regressor for each patch crowd count. Intuitively, it looks into each patch in detail by estimating the count on smaller zoomed-in highly crowded patches. In this way, it greatly stabilizes and improves the count estimate for high-density images, where other methods may either underestimate or overestimate too much due to fixed patch sizes, as demonstrated in the experiments Sec. \ref{experiments}.

\begin{figure}[t]

	\begin{minipage}[b]{\columnwidth}
		\begin{center}
			\centerline{\includegraphics[width=0.995\columnwidth]{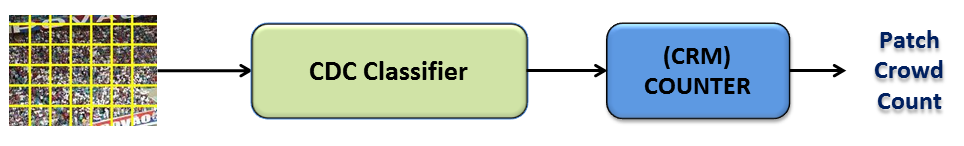}}
		\end{center}
	\end{minipage}		
    \vspace{-7mm}
	\caption{Workflow in case if the patch maker $Z_{out}$ is selected by the decision module for the test image count estimate. The input image is divided into $448\times448$ patches, then down-scaled by $2$ times and fed to the CDC classifier to eliminate no-crowd patches. Crowd patches are then routed to the $COUNTER$ for each patch crowd estimate.}
    \vspace{-3mm}
    \label{fig:flow448}
\end{figure}

\textbf{Zoom-out based Patch Maker} $(Z_{out})$: This block is responsible for handling the low-density extreme case images as detected and routed by the decision module. $Z_{out}$ takes $448\times448$ original patches of the test image $X$, down-scales them by $2$ times, and feeds each resultant patch to the CDC classifier to eliminate any no-crowd patches, as shown in Fig. \ref{fig:flow448}. The count estimate for each crowd patch is then computed through CRM count regressor $(COUNTER)$ followed by the image total count estimate, which is the sum of all patches crowd counts. In other words, it assists the count regressor by using larger area per input patch ($448\times448$ down-scaled to $224\times224$) which alleviates the overestimation problem.

\textbf{Normal case:} In $Normal$ case, the images are divided into $224\times224$ size patches with no up- or down-scaling before patch-wise count regression. It is also worth mentioning that there is no need to explicitly look for and eliminate any no-crowd patches in case of $Normal$ and $Z_{in}$ case images as such background patches are automatically removed during the CDC module classification process, and thus we can also reuse the remaining CDC module crowd patches in both cases for crowd estimate.

The input image that is classified as $Normal$ case, may contain mixed crowd numbers in different regions. Empirically, it has been observed that the deep CNN based crowd counter (i.e. the CRM regressor) can directly handle such images effectively without any rescaling or classification process. Thus, this work only focuses on the images of extreme case (low-density or high-density) that contain most of the regions with same crowd level (low- or high-density), since these extremes have huge influence on crowd count, and a special attention to these cases will significantly mitigate the over- or under-estimation issue as discussed in the introduction.

\subsection{ZoomCount and Switch-CNN Comparison}
Switch-CNN \cite{sam2017switching} is one of the state-of-the-art crowd counting models, and it classifies the input image each patch into one of three crowd-density labels (low, medium, and high). However, our approach employs this classification process fundamentally different from Switch-CNN. The key differences and comparison are as follows:
\noindent \begin{itemize}
\item \textbf{Crowd-Density Classification Purpose and Usage.} Switch-CNN primarily uses the crowd-density classification to route the input patch to one of three specialized count regressors. Subsequently, crowd estimation for each patch is being done using one of these deep count regressors. On the other hand, we employ the crowd-density classification process to facilitate the Decision Module (DM) in detecting whether the input image is either one of the two extremes (low- or high-density) or a normal case image. Then, the patches are rescaled accordingly before proceeding to the only count regressor for the crowd estimation. Thus, the underlying usage and purpose of crowd classification completely differ in both methods.
\item \textbf{Patch-based vs Image-based Decision.} Based on the crowd-density classification as discussed in the above point, Switch-CNN selects the appropriate crowd count regressor individually for each input patch (patch-wise decision). While \textit{ZoomCount} selects the most appropriate rescaling operation suitable for the whole image (image-based decision) using this information. Thus, our method also takes the global context of the image into consideration during the decision-making process instead of relying solely on the local image patches.
\end{itemize}
In addition, Switch-CNN \textbf{trains} using multiple complex training steps including \textit{pretraining, differential training, switch training, and coupled training}, while ZoomCount employs standard classifier and regressor training process. Switch-CNN uses three specialized count regressors, whereas the proposed approach simplifies this process by employing only a single count regressor. Moreover, our method \textbf{outperforms} the Switch-CNN under all evaluation metrics as demonstrated in the experiments section.

\section{Implementation Details}
\label{implementationDetails}
\subsection{Training Details}
\label{Train_Details}
The CDC classifier and the count regressor ($COUNTER$) expect fixed size patch of $224\times224$ as the input. For both modules, we randomly extract $112\times112$, $224\times224$ and $448\times448$ patches from the training images. Around 90,000 such patches with mixed crowd numbers are generated for each of these modules. The count regressor is trained for 80 epochs with Adam optimizer and a batch size of 16 and starting learning rate of 0.001, decreased by half after every 20 epochs. The classifier employs the stochastic gradient descent (SGD) based optimization with multi-step learning rate starting at 0.1 and decreased by half after $25\%$ and $50\%$ epochs with 80 epochs in total. For each dataset, around $10\%$ training data has been used for validation as recommended in the corresponding literature. For the random forest algorithm in RFDB, we utilize machine learning library scikit-learn for python programming. The Random Forest model was trained using 100 RF decision trees, where each RF tree is trained using the bootstrapped sample with {\it Gini Imprity} as node split quality criterion. 10\% of the training data has been used for validation in case of each RFDB dataset.

\subsection{Evaluation Details}

\setlength{\tabcolsep}{1.0pt}
\begin{table}[t]\small
	\caption{\footnotesize Benchmark datasets (used in the experiments) statistics.} 
	\begin{center}
	\begin{tabular}{|c|c|c|c|c|c|}
    \hline
Dataset & Images & Annotations & Min & Max. & Avg.\\ \hline
UCF-QNRF \cite{idrees2018composition} & 1535 & 1,251,642 & 65 & 12865 & 815 \\ \hline
ShanghaiTech Part-A \cite{zhang2016single} & 482 & 241,677 & 33 & 3139 & 501  \\ \hline
WorldExpo'10 \cite{zhang2015cross} & 3980 & 225,216 & 1 & 334 & 56 \\ \hline
AHU-Crowd \cite{hu2016dense} & 107 & 45,807 & 58 & 2201 &  428    \\ \hline

	\end{tabular}
	\end{center}
	\label{table:datasetsComparison}
    \vspace{0mm}
\end{table}

\setlength{\tabcolsep}{2.0pt}
\begin{table}[t]\small
	\caption{\footnotesize Comparison of ZoomCount with the state-of-the-art methods on UCF-QNRF \cite{idrees2018composition} dataset. Methods with '*' do not use density maps at all. Both versions of our method outperform the state-of-the-art on most of the evaluation criteria.}
	\begin{center}
	\begin{tabular}{|c|c|c|c|c|}
    \hline
& MAE & MNAE & RMSE\\ \hline
Idrees et al. \cite{idrees2013multi}* & 315 & 0.63 & 508    \\ \hline
MCNN \cite{zhang2016single}  & 277 & 0.55 & 426    \\ \hline
Encoder-Decoder \cite{badrinarayanan2015segnet} & 270 & 0.56 & 478    \\ \hline
CMTL \cite{cascadedmtl} & 252 & 0.54 & 514    \\ \hline
SwitchCNN \cite{sam2017switching} & 228 & 0.44 & 445    \\ \hline
Resnet101 \cite{he2016deep}* & 190 & 0.50 & 277    \\ \hline
Densenet201\cite{huang2017densely}* & 163 & 0.40 & 226    \\ \hline
CL \cite{idrees2018composition} &  132 & 0.26 & \textbf{191}    \\ \hline \hline
\textbf{ZoomCount-RSE}* & 130 & 0.23 & 204    \\ \hline
\textbf{ZoomCount-RFDB}* & \textbf{128} & \textbf{0.20} & 201    \\ \hline
	\end{tabular}
	\end{center}
	\label{table:ucf_results}
    \vspace{0mm}
\end{table}

\setlength{\tabcolsep}{2.0pt}
\begin{table}[t]\small
	\caption{{\footnotesize Ablation experiments on UCF-QNRF \cite{idrees2018composition} dataset emphasize importance of zoom-in, zoom-out patch-making blocks and associated rules in ZoomCount-RSE. The first eight rows depict the effect of removing one rule at a time on MAE, MNAE and RMSE while next three rows demonstrate the effect without using the zoom-in ($Z_{in}$), zoom-out ($Z_{out}$) and both zoom-in and zoom-out blocks respectively, followed by the method with original setting in last row. $I_{Zin},I_N,I_{Zout}$ indicate the total images handled by zoom-in, normal and zoom-out patch-makers respectively before proceeding to the count regressor.}}
	\begin{center}
	\begin{tabular}{|c|c|c|c|c|c|c|c|}
    \hline
Without & MAE & MNAE & RMSE & $I_{Zin}$ & $I_N$ & $I_{Zout}$\\ \hline
R1 & 133.1 & 0.24 & 204 & 75 & 128 & 131   \\ \hline
R2 & 135.7 & 0.26 & 224 & 75 & 140 & 119   \\ \hline
R3 & 138.4 & 0.27 & 230 & 75 & 150 & 109   \\ \hline
R4 & 143.7 & 0.27 & 236  & 75 & 132 & 127  \\ \hline
R5 & 134.7 & 0.24 & 215 & 67 & 105 & 162\\ \hline
R6 & 132.6 & 0.23 & 211 & 68 & 104 & 162\\ \hline
R7 & 131.9 & 0.23 & 209 & 72 & 105 & 157\\ \hline
R8 & 140.7 & 0.27 & 219  & 18 & 154 & 162  \\ \hline \hline
$Z_{in}$ & 141.3 & 0.27 & 220   & 0 & 172 & 162 \\ \hline
$Z_{out}$ & 150.0 & 0.31 & 244  & 75 & 259 & 0  \\ \hline
$Z_{in}$ \&$Z_{out}$ & 163.0 & 0.40 & 226  & 0 & 334 & 0  \\ \hline
- & \textbf{130} & \textbf{0.23} & \textbf{204} & 75 & 97 & 162\\ \hline
	\end{tabular}
	\end{center}
	\label{table:ablation_UCF}
    \vspace{0mm}
\end{table}

In order to make a fair and consistent comparison with other methods, we employ three evaluation metrics namely Mean Absolute Error ($MAE$), Mean Normalized Absolute Error ($MNAE$) and Root Mean Squared Error ($RMSE$) defined as below:
\begin{equation}
\label{eq3}
MAE = \frac{1}{N} \sum_{i=1}^{N} |Y_{i}-\hat{Y_{i}}|
\end{equation}
\begin{equation}
\label{eq4}
MNAE = \frac{1}{N} \sum_{i=1}^{N} \frac{|Y_{i}-\hat{Y_{i}}|}{Y_{i}}
\end{equation}
\begin{equation}
\label{eq5}
RMSE =\sqrt[]{ \frac{1}{N} \sum_{i=1}^{N} (Y_{i}-\hat{Y_{i}})^{2}}
\end{equation}
where $N$ denotes the total number of test images, and $Y_{i}\ and\ \hat{Y_{i}}$ are the ground truth and the estimated counts respectively for the test image $i$.

\begin{figure}[t]

	\begin{minipage}[b]{\columnwidth}
		\begin{center}
			\centerline{\includegraphics[width=0.995\columnwidth]{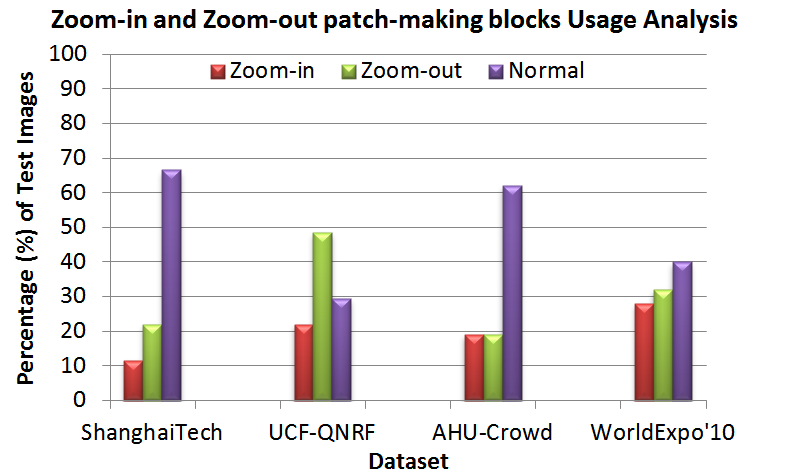}}
		\end{center}
	\end{minipage}		
    \vspace{-10mm}
	\caption{Quantitative importance of zoom-in and zoom-out blocks and given rule sets. For each benchmark, at least $11\%$ and as high as $48.5\%$ test images pass through one of these specialized patch-makers before patch-wise count regression, demonstrating the value and effectiveness of such blocks and associated rules. }
    \vspace{-0mm}
    \label{fig:countersImportance}
\end{figure}

\section{Experiments}
\label{experiments}

In this section, we demonstrate both quantitative and qualitative results from extensive four benchmark datasets: UCF-QNRF \cite{idrees2018composition}, ShanghaiTech \cite{zhang2016single}, WorldExpo'10 \cite{zhang2015cross}, and AHU-Crowd \cite{hu2016dense}. These datasets contain images with huge crowd variance, different camera perspective, and complex cluttered background regions. Details about each benchmark are given in Table \ref{table:datasetsComparison}. At the end of this section, we also discuss computational time analysis and compare that with state-of-the-art methods.

Two different versions of the proposed model, the Rule-Set Engine  (\textbf{ZoomCount-RSE}) based and the automated RFDB module (\textbf{ZoomCount-RFDB}) based version, are being compared separately with the state-of-the-art techniques throughout this section. Both ZoomCount versions give almost identical and much better performance under most of the evaluation criteria on the four benchmark datasets.
\subsection{Experiments on UCF-QNRF Dataset}
The dataset was recently published by Idrees {\it et al.} \cite{idrees2018composition}, which is a challenging and the first dataset of its kind. On one hand, it contains images with resolution as high as $(6666\times9999)$ and as low as $(300\times377)$; on the other hand, crowd count per image ranges from a maximum value of $12,865$ to a minimum count of $65$. The total number of annotations in this dataset is $1,251,642$, indicating the level of crowd complexity. It contains 1535 images in total, out of which 1201 and 334 images are used for training and testing respectively. We compare ZoomCount with the state-of-the-art methods and tabulate the results in Table \ref{table:ucf_results}. It is evident that both versions of our method outperform all other approaches in terms of MAE and MNAE; while performing competitively closer to the best in terms of RMSE.

 In order to evaluate the influence of different rules, we perform the ablation experiments, as shown in Table \ref{table:ablation_UCF}. We analyze the effect of all rules (R1 to R8) by removing them one at a time in ZoomCount-RSE version of the proposed method. As shown in the results, doing so greatly decreases the performance of our method, thus demonstrating the importance of those rules. We also analyze the effect of removing both or either of the zoom-in and zoom-out patch-makers in the experiments. From the results in Table \ref{table:ablation_UCF}, it is evident that both modules play an effective role in improving the overall performance of our method. The last three columns of Table \ref{table:ablation_UCF} show the number of the test images passed through the zoom-in, normal and zoom-out patch-makers respectively. In the original setting, $75$ ($\sim22\%$) images passed through the zoom-in patch-maker, whereas the zoom-out block handled $162$ ($\sim48.5\%$) images and normal patch-maker was used only for $97$ ($\sim29.5\%$) images, showcasing quantitative importance of these extreme case handlers, as shown in Fig. \ref{fig:countersImportance}. We also compare the crowd estimate of ten test images each, for both extreme cases with DenseNet\cite{huang2017densely} direct regression and the state-of-the-art CL \cite{idrees2018composition}  density map method. Our method performs much better in both cases, as shown in Fig. \ref{fig:graphi_ten_cases}.

\setlength{\tabcolsep}{2.0pt}
\begin{table}[t]\small
	\caption{\footnotesize Comparison of ZoomCount with the state-of-the-art approaches on the ShanghaiTech \cite{zhang2016single} dataset, where '*' indicates methods not using density maps at all. Our method performs the best on every evaluation criteria.}
	\begin{center}
	\begin{tabular}{|c|c|c|c|c|}
    \hline
& MAE & MNAE & RMSE\\ \hline
Zhang et al. \cite{zhang2015cross} & 181.8  & - & 277.7    \\ \hline
MCNN \cite{zhang2016single}  & 110.2 & - & 173.2    \\ \hline
Cascaded-MTL \cite{cascadedmtl} & 101.3 & 0.279 & 152.4    \\ \hline
Switch-CNN \cite{sam2017switching} & 90.4 & - & 135.0    \\ \hline
CP-CNN \cite{sindagi2017generating} & 73.6 & - & 106.4    \\ \hline
CSRNet \cite{li2018csrnet} & 68.2 & - & 115.0    \\ \hline

IG-CNN \cite{babu2018divide} & 72.5 & - & 118.2    \\ \hline
L2R \cite{liu2018leveraging} & 72.0 & - & 106.6    \\ \hline
ICC \cite{ranjan2018iterative} & 68.5 & - & 116.2    \\ \hline
SA-Net \cite{cao2018scale} &  67.0 & - &  104.5    \\ \hline
Deep-NCL \cite{shi2018crowd} & 73.5 & - & 112.3    \\ \hline

Densenet201\cite{huang2017densely}* & 79.3 &  0.224 & 118.9    \\ \hline \hline
\textbf{ZoomCount-RSE}* & 66.6 & 0.197 & \textbf{94.5}    \\ \hline
\textbf{ZoomCount-RFDB}* & \textbf{66.0} & \textbf{0.190} & 97.5    \\ \hline
	\end{tabular}
	\end{center}
	\label{table:ST_results}
    \vspace{-2mm}
\end{table}

\setlength{\tabcolsep}{2.0pt}
\begin{table}[t]\small
	\caption{{\footnotesize Ablation experiments on the ShanghaiTech \cite{zhang2016single} dataset show quantitative importance of the zoom-in, zoom-out patch-making blocks and associated rules in ZoomCount-RSE. The first eight rows depict the effect of removing one rule at a time. Next three rows demonstrate the results without using the zoom-in ($Z_{in}$), zoom-out ($Z_{out}$) and both zoom-in and zoom-out blocks respectively, followed by the method with original setting in the last row. $I_{Zin},I_N,I_{Zout}$ indicate the total images handled by zoom-in, normal and zoom-out blocks respectively before proceeding to the counter.}}
	\begin{center}
	\begin{tabular}{|c|c|c|c|c|c|c|}
    \hline
Without & MAE & MNAE & RMSE & $I_{Zin}$ & $I_N$ & $I_{Zout}$\\ \hline
R1 & 66.8 & 0.199 & 95.3 & 21 & 135 & 26   \\ \hline
R2 & 66.8 & 0.198 & 96.0 & 21 & 122 & 39   \\ \hline
R3 & 67.2 & 0.198 & 94.7 & 21 & 124 & 37   \\ \hline
R4 & 68.0 & 0.206 & 95.9 & 21 & 143 & 18  \\ \hline

R5 & 69.4 & 0.200 & 103.7 & 15 & 127 & 40\\ \hline
R6 & 69.1 & 0.210 & 101.8 & 19 & 123 & 40\\ \hline
R7 & 66.8 & 0.200 & 97.2 & 20 & 122 & 40\\ \hline
R8 & 69.4 & 0.199 & 97.2  & 09 & 133 & 40  \\ \hline \hline

$Z_{in}$ & 74.9 & 0.200 & 116.4   & 0 & 142 & 40 \\ \hline
$Z_{out}$ & 69.1 & 0.210 & 96.5  & 21 & 161 & 0  \\ \hline
$Z_{in}$ \& $Z_{out}$ & 79.3 & 0.224 & 118.9  & 0 & 182 & 0  \\ \hline
- & \textbf{66.6} & \textbf{0.197} & \textbf{94.5} & 21 & 121 & 40\\ \hline
	\end{tabular}
	\end{center} 
	\label{table:ablation_ST}
    \vspace{1mm}
\end{table}

\begin{figure}[t]
	\begin{minipage}[t]{\columnwidth}
		\begin{center}
			\centerline{\includegraphics[width=0.995\columnwidth]{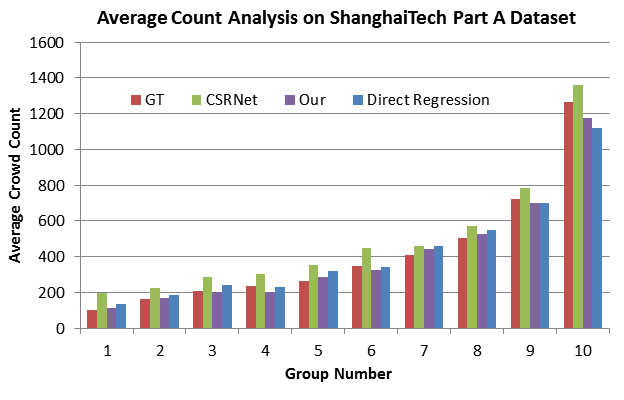}}
		\end{center}
	\end{minipage}		
    \vspace{-5mm}
	\caption{182 test images are divided into ten groups with total crowd count in each group increasing from left to right. Each group contains 18 images except group number 10. Vertical axis indicates average count for each group. It is evident that ZoomCount remains closer to the ground truth (GT) bar in most cases as compared to the state-of-the-art methods.}
    \vspace{-2mm}
    \label{fig:st_histogram}
\end{figure}

\setlength{\tabcolsep}{2.0pt}
\begin{table}[t]\small
	\caption{\footnotesize MAE metric based ZoomCount scheme comparison with the state-of-the-arts on the WorldExpo'10 \cite{zhang2015cross} dataset on five test scenes (S1-S5), where '*' indicates methods without using density maps. Our method outperforms previous approaches on two test scenes (S1 and S3) and the average MAE metric, while perform reasonably closer to the best for other test scenes.}
	\begin{center}
	\begin{tabular}{|c|c|c|c|c|c|c|}
    \hline
 & S1 & S2 & S3 & S4 & S5 & Avg. \\ \hline
Zhang et al. \cite{zhang2015cross} & 9.8 & 14.1 & 14.3 & 22.2 & \textbf{3.7} & 12.9   \\ \hline
MCNN \cite{zhang2016single} & 3.4 & 20.6 & 12.9 & 13.0 & 8.1 & 11.6   \\ \hline
Switch-CNN \cite{sam2017switching} & 4.4 & 15.7 & 10.0 & 11.0 & 5.9 & 9.4   \\ \hline
CP-CNN \cite{sindagi2017generating} & 2.9 & 14.7 & 10.5 & 10.4 & 5.8 & 8.9  \\ \hline 
IG-CNN \cite{babu2018divide} & 2.6 & 16.1 & 10.15 & 20.2 & 7.6 & 11.3 \\ \hline
IC-CNN \cite{ranjan2018iterative} & 17.0 & \textbf{12.3} & 9.2 & \textbf{8.1} & 4.7 & 10.3 \\ \hline
Densenet201\cite{huang2017densely}* & 4.3 & 17.9 & 12.7  & 13.1 & 6.9 & 11 \\ \hline 
\textbf{ZoomCount-RSE}* & 2.3 & 16.2 & 10.4  & 9.7 & 4.8 & 8.7  \\ \hline 
\textbf{ZoomCount-RFDB}* & \textbf{2.1} & 15.3 & \textbf{9.0}  & 10.3 & 4.5 & \textbf{8.3}  \\ \hline 
	\end{tabular}
	\end{center} 
	\label{table:table_we10}
    \vspace{-5mm}
\end{table}

\setlength{\tabcolsep}{2.0pt}
\begin{table}[t]\small
	\caption{\footnotesize Comparison of ZoomCount with the state-of-the-art on the AHU-Crowd \cite{hu2016dense} dataset, where '*' indicates methods without using density maps. Our method outperforms previous approaches on all evaluation metrics.}
	\begin{center}
	\begin{tabular}{|c|c|c|c|c|}
    \hline
& MAE & MNAE & RMSE\\ \hline
Haar Wavelet \cite{oren1997pedestrian} & 409.0  & 0.912 & -    \\ \hline
DPM \cite{felzenszwalb2008discriminatively} & 395.4 & 0.864 & -    \\ \hline
BOW–SVM \cite{csurka2004visual} & 218.8 & 0.604 & -    \\ \hline
Ridge Regression \cite{chen2012feature} & 207.4 & 0.578 & -    \\ \hline
Hu et al. \cite{hu2016dense} & 137 & 0.365 & -    \\ \hline

DSRM \cite{yao2017deep} &  81 &  0.199 &  129    \\ \hline

Densenet201\cite{huang2017densely}* & 87.6 & 0.295 & 124.9    \\ \hline \hline
\textbf{ZoomCount-RSE}* & 79.3 & 0.198 & 121    \\ \hline
\textbf{ZoomCount-RFDB}* & \textbf{74.9} & \textbf{0.190} & \textbf{111}    \\ \hline
	\end{tabular}
	\end{center}
	\label{table:AHU_results}
    \vspace{-5mm}
\end{table}

\setlength{\tabcolsep}{2.0pt}
\begin{table}[t]\small
	\caption{\footnotesize Comparison of different state-of-the-art architectures for the CDC classifier and the CRM Regressor choice on the ShanhaiTech Part-A dataset \cite{zhang2016single}. CDC classifier is evaluated using the classification accuracy metric, while CRM regressor choice experiments are conducted using the mean absolute error (MAE) metric. For the regression experiments, the architectures are evaluated separately in two different settings; as the CRM regressor as well as the standalone regression based crowd count model. DenseNet-201 appears to be the best choice for both modules.}
	\begin{center}
	\begin{tabular}{|c|c|c|c|c|}
    \hline
Model &  \pbox{22cm}{\vspace{.3\baselineskip} \relax\ifvmode\centering\fi CDC \\ Class. \\ Accuracy}  & \pbox{22cm}{ \vspace{.3\baselineskip} \relax\ifvmode\centering\fi
CRM Regressor \\ (MAE)} & \pbox{22cm}{ \vspace{.3\baselineskip} \relax\ifvmode\centering\fi Standalone \\ Crowd Count CRM \\ Regressor (MAE)} \\ \hline
VGG-16 \cite{simonyan2014very} & 74.9  & 95.2 & 109.4    \\ \hline
ResNet-101 \cite{he2016deep} & 86.6 & 80.7 & 90.3    \\ \hline
DenseNet-201 \cite{huang2017densely} & \textbf{93.2} & \textbf{66.0} & \textbf{79.3}    \\ \hline
	\end{tabular}
	\end{center}
	\label{table:choice_class_reg_results}
    \vspace{-6mm}
\end{table}

\subsection{Experiments on ShanghaiTech Dataset}
The ShanghaiTech part A dataset contains a total of 482 images with 241,677 annotations, randomly collected from the internet, with a split of 300 and 182 images for training and testing respectively. We compare our method with the state-of-the-art methods as shown in Table \ref{table:ST_results}. The results show that our method outperforms all other methods on every evaluation metric with significant improvement from $0.224$ to $0.190 (\sim15\%)$ in terms of MNAE and from $104.5$ to $94.5 (\sim9.6\%)$ in case of RMSE.

The proposed rules (R1-R8) play an important and effective role in the performance improvement of ZoomCount-RSE version of our method as shown in Table \ref{table:ablation_ST}, where we remove each rule one at a time. It is clear that the error increases by removing these rules. In the same table, We also analyze the effect of removing the zoom-in and zoom-out blocks separately and together. As expected, the performance plunges dramatically as error increases without using them. The last three columns show the number of test images passing through the zoom-in, normal and zoom-out patch-makers respectively. In the original setting, $21$ ($\sim11.5\%$), $121$ ($\sim66.5\%$) and $40$ ($\sim22\%$) images are handled by the zoom-in, normal and zoom-out blocks respectively, thus proving the quantitative importance of all of them and associated rules in ZoomCount-RSE, as shown in Fig. \ref{fig:countersImportance}.
In Fig. \ref{fig:st_histogram}, we analyze the performance of our method on the average count across image groups with different total crowd counts. As compared with the state-of-the-art methods, ZoomCount performs the best in most cases.

\subsection{Experiments on WorldExpo'10 Dataset}
The WorldExpo'10 \cite{zhang2015cross} is a large dataset, composed of 1132 video sequences taken by 108 different cameras. The training set consists of 3380 images from 103 different scenes, whereas the testing set has 5 scenes with a total of 600 frames. This benchmark also consists of Region of Interest (RoI) and perspective maps. We only utilize the RoIs in the images during training and testing stages. The MAE based evaluation results on the five test scenes and the average MAE error are shown in Table \ref{table:table_we10}. As shown, the proposed model achieves the lowest average MAE and the best performance on two scenes (S1 and S3).  For the other three scenes, the proposed scheme yields reasonable and competitive results to the state-of-the-art. During the testing experiments on ZoomCount-RSE model, $168$ ($\sim28\%$), $240$ ($\sim40\%$) and $192$ ($\sim32\%$) images are handled by the zoom-in, normal, and zoom-out blocks, respectively, as shown in Fig. \ref{fig:countersImportance}. This demonstrates the quantitative importance of the associated rules in ZoomCount-RSE.
\subsection{Experiments on AHU-Crowd Dataset}
AHU-Crowd \cite{hu2016dense} dataset contains 107 images with $45,807$ human annotations. The crowd count ranges from $58$ to $2201$ per image. As per the standard being followed for this dataset \cite{hu2016dense}, we performed 5-fold cross-validation and evaluated our method using the same three evaluation metrics. ZoomCount outperforms all the other methods as shown in Table \ref{table:AHU_results}. It is worth-mentioning that ZoomCount decreases MAE and MNAE significantly by $\sim7.5\%$ ($81$ to $74.9$) and $\sim4.5\%$ ($0.199$ to $0.190$) respectively, whereas RMSE decreases drastically by $\sim11.2\%$ ($124.9$ to $111$).

\setlength{\tabcolsep}{2.0pt}
\begin{table}[t]\small
	\caption{\footnotesize The classification accuracy results of the RSE and the RFDB based versions of the DM module on the four benchmarks. These results directly effect the overall performance of the proposed framework.}
	\begin{center}
	\begin{tabular}{|c|c|c|c|c|}
    \hline
Benchmark &  RSE Accuracy(\%)  & RFDB Accuracy (\%) \\ \hline
UCF-QNRF \cite{idrees2018composition} & 87.8  & 93.0     \\ \hline
ShanghaiTech \cite{zhang2016single} & 92.0 & 91.4     \\ \hline
WorldExpo'10 \cite{zhang2015cross} & 91.7 & 94.2     \\ \hline
AHU-Crowd \cite{hu2016dense} \cite{zhang2015cross} & 89.2 & 92.3     \\ \hline
	\end{tabular}
	\end{center}
	\label{table:dm_acc_results}
    \vspace{-5mm}
\end{table}

\subsection{CDC Classifier and CRM Regressor Architecture Selection}
Choosing an appropriate network architecture for the CDC classifier and the CRM regressor is essential for the effectiveness of the proposed scheme. In this section, we analyze different state-of-the-art architectures for this objective, including VGG-16 \cite{simonyan2014very}, ResNet-101 \cite{he2016deep}, and DenseNet-201 \cite{huang2017densely}. For the classifier choice evaluation, the final 1000-way classification layer in VGG-16 and ResNet-101 is replaced with a 4-way classification layer. Similarly, for the CRM regressor evaluation, the final FC layer in VGG-16 and ResNet-101 networks is followed by a single neuron to directly regress the crowd count with mean square error (MSE) loss function as given in Eq. \ref{eq2}. As per the standard practice, VGG-16 is configured with the batch-normalization layer being added in-between each convolution and activation layer. DenseNet-201 classifier and regressor are configured as stated in sec. \ref{IPCM}, and all architectures are trained as discussed in section \ref{Train_Details}.

The results on the test images of ShanghaiTech Part-A dataset are shown in Table \ref{table:choice_class_reg_results}. We can see from the results that DenseNet-201 based classifier outperforms other state-of-the-art architectures with the highest classification accuracy and $\sim 7\%$ improvement, thus, we choose this model as the CDC classifier in the proposed method. For the choice of the CRM Regressor, we evaluated different architectures in terms of mean absolute error (MAE) in two separate experimental settings. In the first case, each regressor architecture has been evaluated separately by plugging-in into the proposed scheme as the CRM Regressor with the same architecture, while in the second experiment setting, each regressor architecture has been evaluated independently as a standalone crowd count regression model. The results of both experimental settings are shown in the second-rightmost and the rightmost columns respectively of Table \ref{table:choice_class_reg_results}. From the results, we can see that using DenseNet based architecture as the CDC classifier and CRM Regressor achieves the highest classification accuracy and the lowest MAE errors, respectively.

\setlength{\tabcolsep}{2.0pt}
\begin{table}[t]\small
	\caption{\footnotesize ZoomCount performance analysis on ShanghaiTech and UCF-QNRF benchmarks using different ML classification algorithms in the RFDB block of Decision Module (DM). As shown, top five results indicate best performance by the Random Forest algorithm, thus, justifying its usage in the RFDB module.} 

	\begin{center}
	\begin{tabular}{|c|c|c|c|c|c|c|}
    \hline
 & \multicolumn{3}{c|}{ShanghaiTech}& \multicolumn{3}{c|}{UCF-QNRF} \\ \cline{2-7}
 & MAE & MNAE & RMSE & MAE & MNAE & RMSE\\ \hline

\textbf{Random Forest} & \textbf{66.0} & \textbf{0.190} & \textbf{97.5} & \textbf{128} & \textbf{0.20} & \textbf{201}   \\ \hline
ExtraTrees & 70.7 & 0.20 & 102.8 & 135 & 0.22 & 214   \\ \hline
GradientBoosting & 72.9 & 0.22 & 119.0 & 137 & 0.22 & 222   \\ \hline
AdaBoost & 75.0 & 0.22 & 105.31 & 151 & 0.24 & 265   \\ \hline
Logistic Regression & 78.9 & 0.23 & 119.6 & 177 & 0.24 & 279   \\ \hline

	\end{tabular}
	\end{center}
	\label{table:ablation_ml}
    \vspace{-4mm}
\end{table}

\subsection{RSE and RFDB based Decision Modules Performance}
The classification accuracy of the decision module directly affects the overall performance of the proposed method. In this section, we show the individual classification accuracy performance of the RSE and RFDB based DM modules on the four benchmarks. The accuracy results are shown in Table \ref{table:dm_acc_results}, from which we can see that both modules perform quite effectively. Also, the RFDB version of the DM module gives slightly better accuracy in most cases compared to the RSE based DM module. The overall system outperforms the state-of-the-art owing to the improved and reasonable classification accuracy of the DM module.

\begin{figure*}

	\begin{minipage}[c]{0.39\columnwidth}
		\begin{center}
			\centerline{\includegraphics[width=1\columnwidth]{figures/fig2/rsz_img_0074.jpg}}
			\centerline{\footnotesize{GT=704, DR=861}}
			\centerline{\footnotesize{Ours=708, Density\cite{ucfonlinedemosite}=1017}}
		\end{center}
	\end{minipage}
		\begin{minipage}[c]{0.39\columnwidth}
		\begin{center}
			\centerline{\includegraphics[width=1\columnwidth]{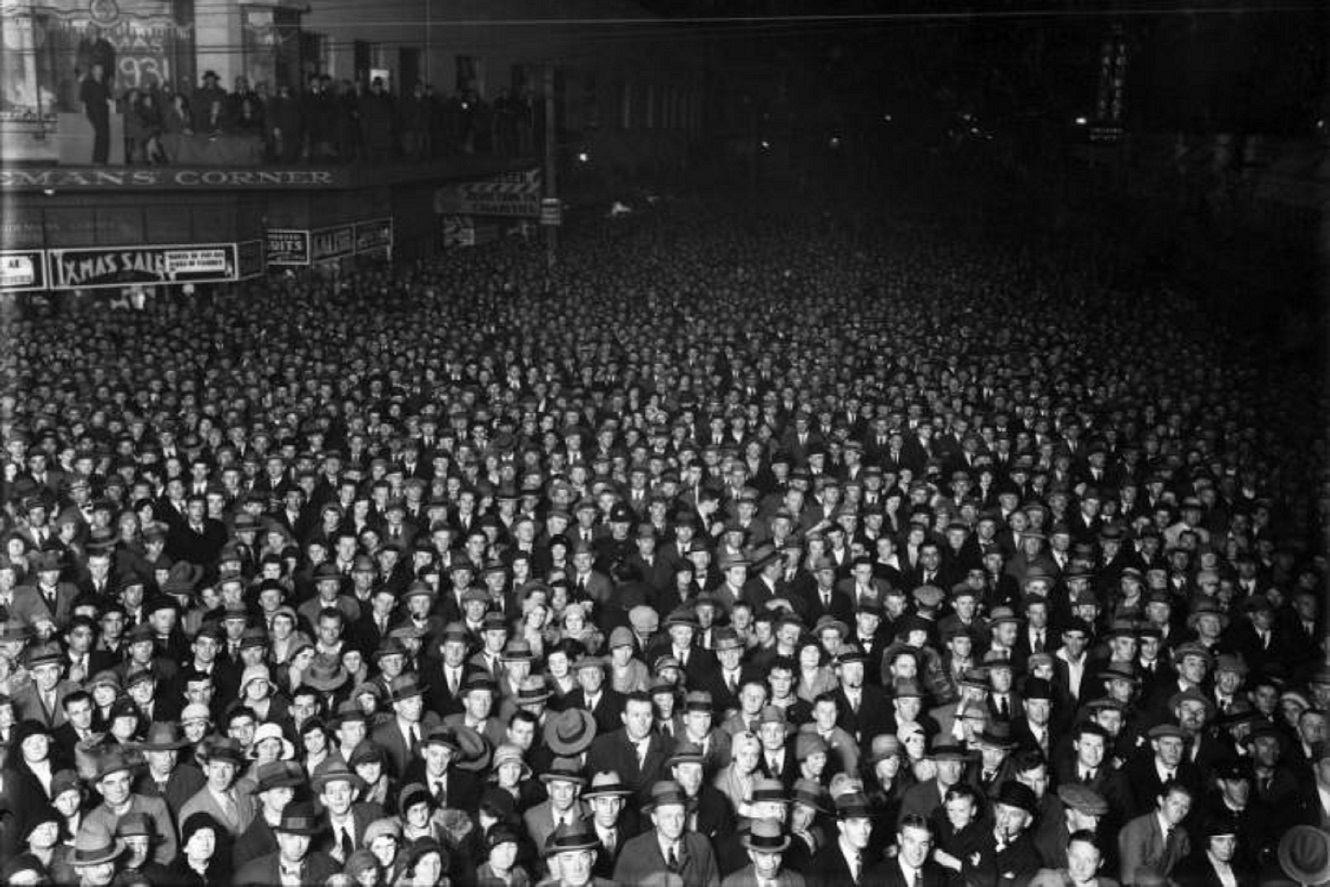}}
			\centerline{\footnotesize{GT=1443, DR=1516}}
			\centerline{\footnotesize{Ours=1443, Density\cite{ucfonlinedemosite}=388}}
		\end{center}
	\end{minipage}
	\begin{minipage}[c]{0.39\columnwidth}
		\begin{center}
			\centerline{\includegraphics[width=1\columnwidth]{figures/fig2/img_0058.jpg}}
			\centerline{\footnotesize{GT=4535, DR=4109}}
			\centerline{\footnotesize{Ours=4523, Density\cite{ucfonlinedemosite}=2759}}
		\end{center}
	\end{minipage}
	\begin{minipage}[c]{0.39\columnwidth}
		\begin{center}
			\centerline{\includegraphics[width=1\columnwidth]{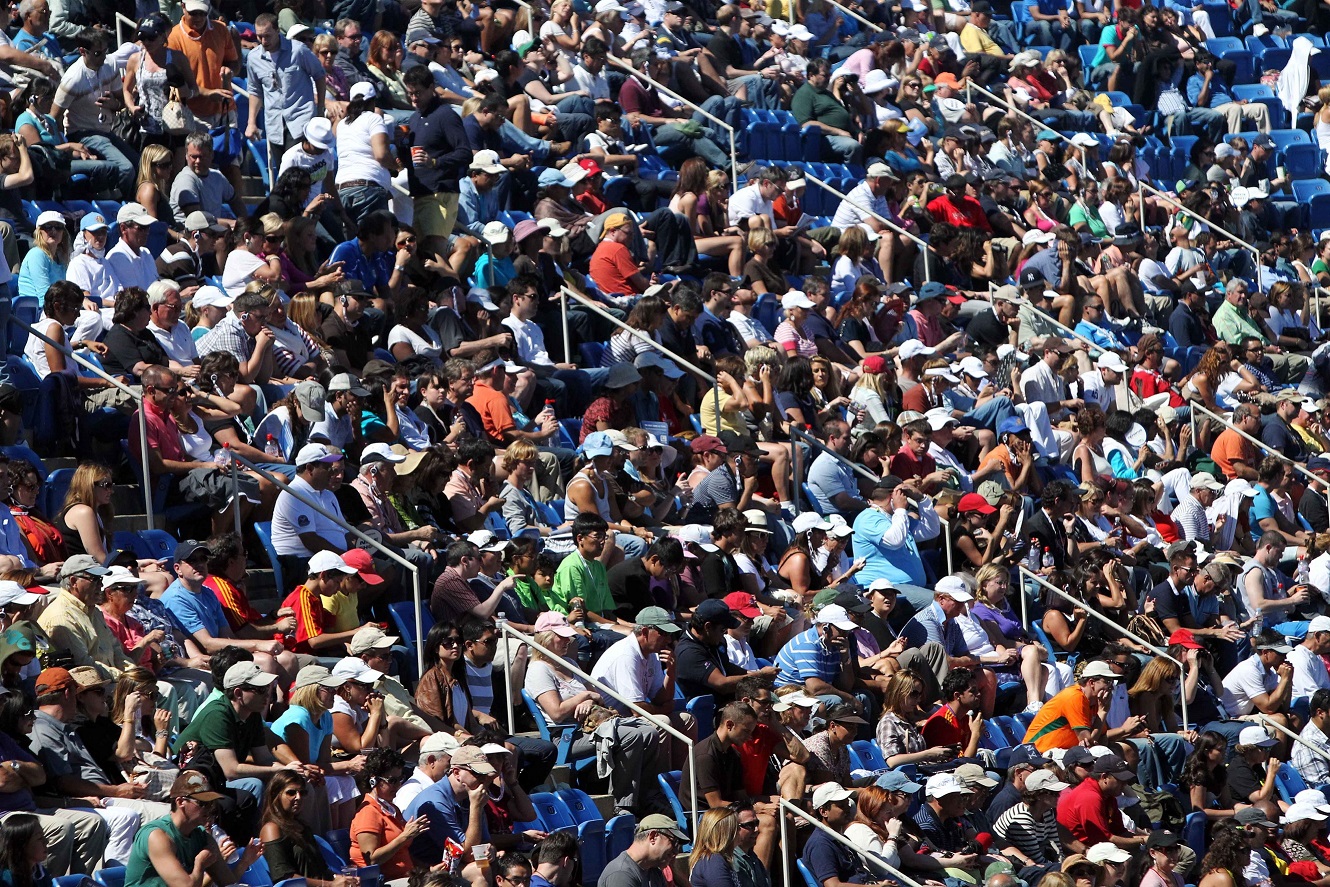}}
			\centerline{\footnotesize{GT=353, DR=493}}
			\centerline{\footnotesize{Ours=489, Density\cite{ucfonlinedemosite}=476}}
		\end{center}
	\end{minipage}			
		\begin{minipage}[c]{0.39\columnwidth}
		\begin{center}
			\centerline{\includegraphics[width=1\columnwidth]{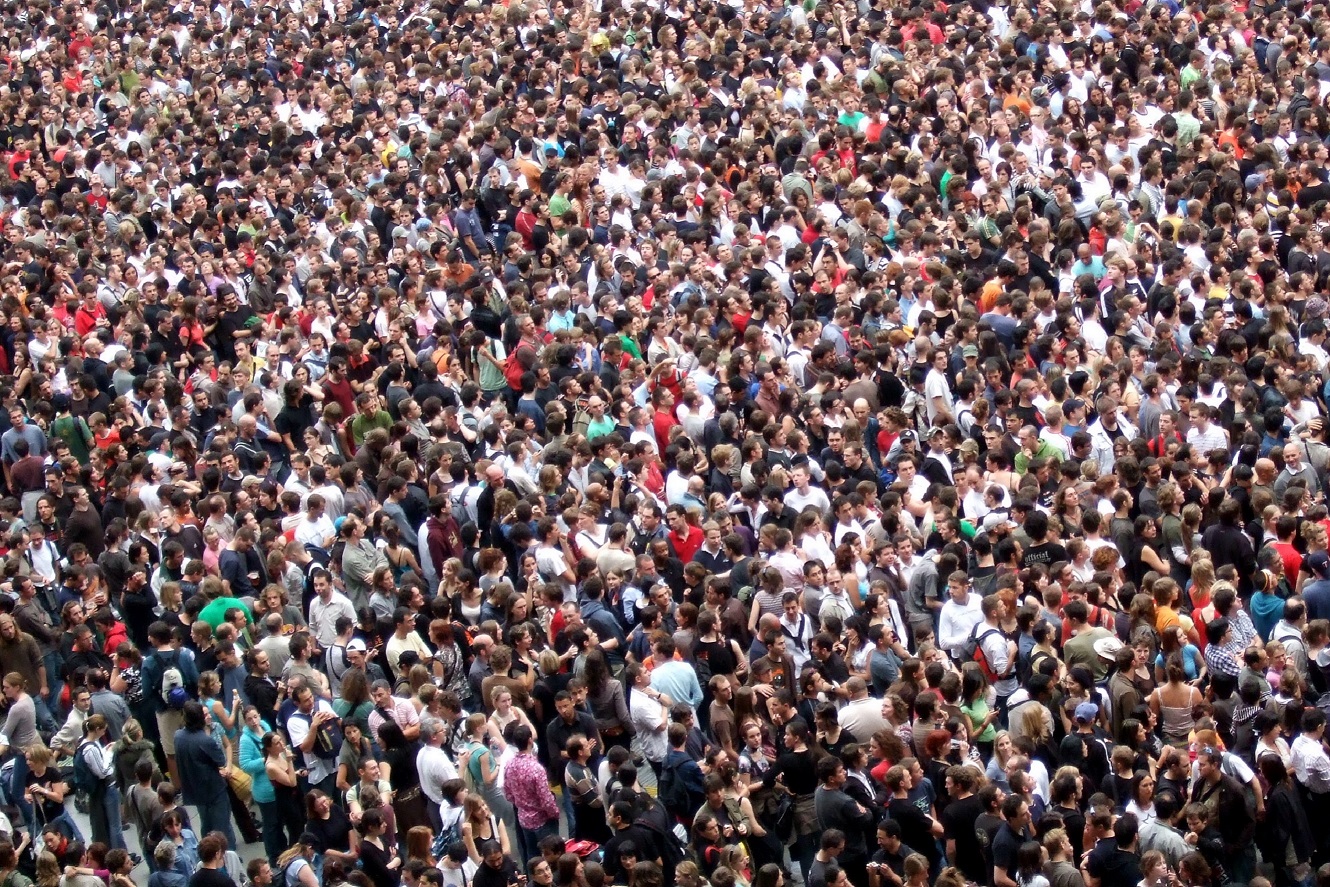}}
			\centerline{\footnotesize{GT=1668, DR=1629}}
			\centerline{\footnotesize{Ours=1476, Density\cite{ucfonlinedemosite}=1717}}
		\end{center}
	\end{minipage}	
	
		\begin{minipage}[c]{0.39\columnwidth}
		\begin{center}
			\centerline{\includegraphics[width=1\columnwidth]{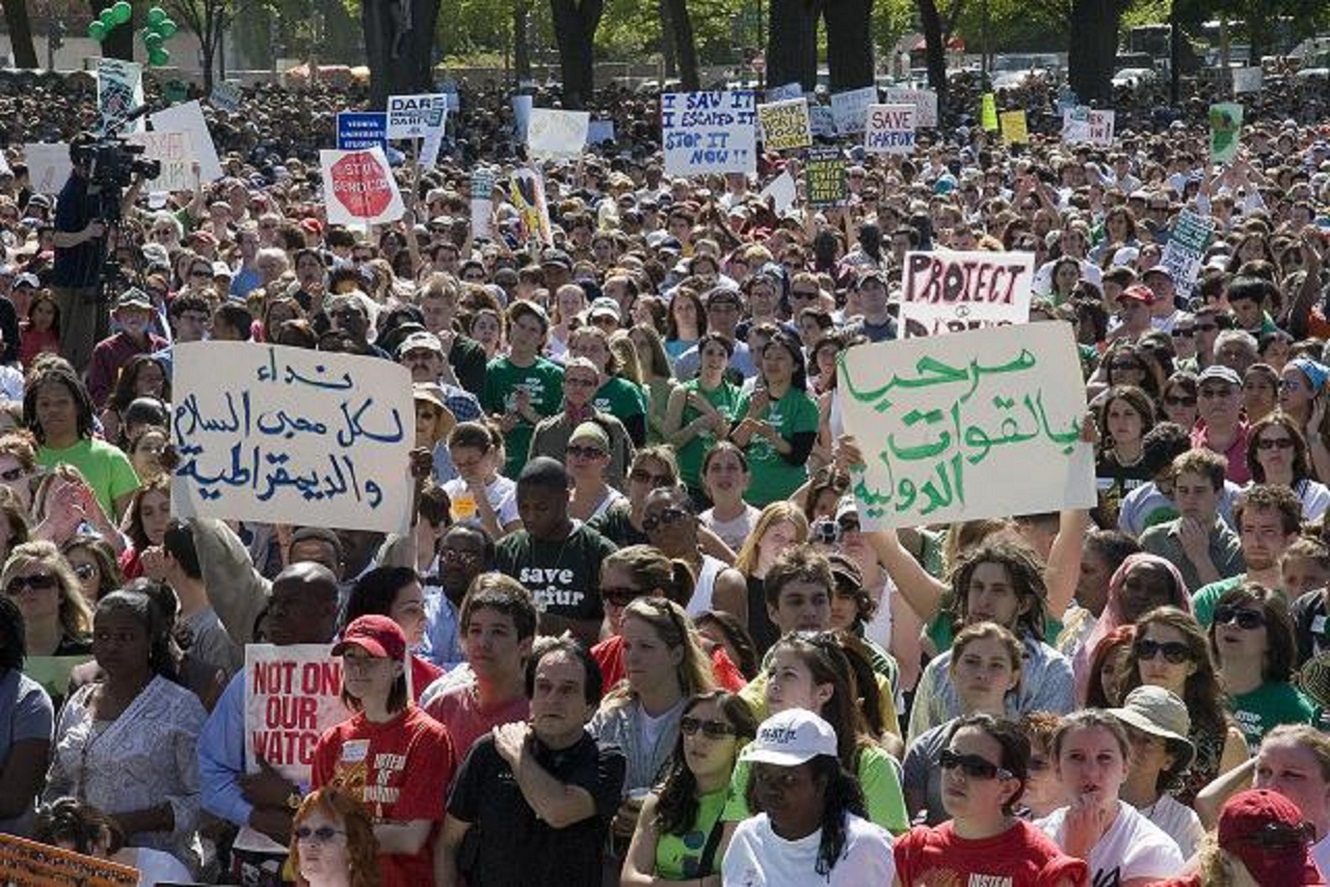}}
			\centerline{\footnotesize{GT=297, DR=474}}
			\centerline{\footnotesize{Ours=299, Density\cite{li2018csrnet}=457}}
		\end{center}
	\end{minipage}
	\begin{minipage}[c]{0.39\columnwidth}
		\begin{center}
			\centerline{\includegraphics[width=1\columnwidth]{figures/fig2/IMG_32.jpg}}
			\centerline{\footnotesize{GT=961, DR=997}}
			\centerline{\footnotesize{Ours=996, Density\cite{li2018csrnet}=1022}}
		\end{center}
	\end{minipage}
		\begin{minipage}[c]{0.39\columnwidth}
		\begin{center}
			\centerline{\includegraphics[width=1\columnwidth]{figures/fig2/IMG_92.jpg}}
			\centerline{\footnotesize{GT=1366, DR=1425}}
			\centerline{\footnotesize{Ours=1384, Density\cite{li2018csrnet}=1445}}
		\end{center}
	\end{minipage}
	\begin{minipage}[c]{0.39\columnwidth}
		\begin{center}
			\centerline{\includegraphics[width=1\columnwidth]{figures/fig2/IMG_41.jpg}}
			\centerline{\footnotesize{GT=249, DR=159}}
			\centerline{\footnotesize{Ours=140, Density\cite{li2018csrnet}=174}}
		\end{center}
	\end{minipage}			
		\begin{minipage}[c]{0.39\columnwidth}
		\begin{center}
			\centerline{\includegraphics[width=1\columnwidth]{figures/fig2/IMG_159.jpg}}
			\centerline{\footnotesize{GT=199, DR=321}}
			\centerline{\footnotesize{Ours=413, Density\cite{li2018csrnet}=413}}
		\end{center}
	\end{minipage}			
    \vspace{-4mm}
	\caption{\footnotesize{Some examples of the qualitative results. First and second rows show qualitative results of our method on the UCF-QNRF \cite{idrees2018composition} and ShanghaiTech \cite{zhang2016single} datasets respectively. First three columns show good results, followed by two bad case images. Each result also shows the estimates of DenseNet \cite{huang2017densely} Direct Regression (DR) and the Density map method as a comparison.
	}}
	\label{fig:qualityResults}
    \vspace{-3mm}
\end{figure*}

\begin{figure}[t]
	\begin{minipage}[b]{0.242\columnwidth}
		\begin{center}
			\centerline{\includegraphics[width=0.995\columnwidth]{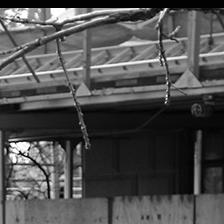}}
		\end{center}
	\end{minipage}
	\begin{minipage}[b]{0.242\columnwidth}
		\begin{center}
			\centerline{\includegraphics[width=0.995\columnwidth]{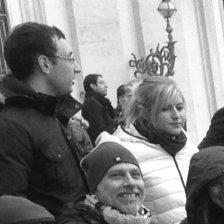}}
		\end{center}
	\end{minipage}
		\begin{minipage}[b]{0.242\columnwidth}
		\begin{center}
			\centerline{\includegraphics[width=0.995\columnwidth]{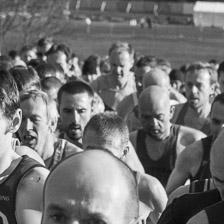}}
		\end{center}
	\end{minipage}
	\begin{minipage}[b]{0.242\columnwidth}
		\begin{center}
			\centerline{\includegraphics[width=0.995\columnwidth]{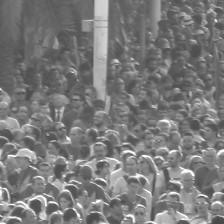}}
		\end{center}
	\end{minipage}
		
			\begin{minipage}[b]{0.242\columnwidth}
		\begin{center}
			\centerline{\includegraphics[width=0.995\columnwidth]{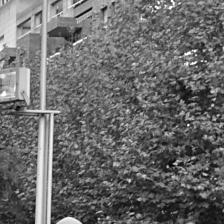}}
			\centerline{\footnotesize{no-crowd(NC)}}
		\end{center}
	\end{minipage}
	\begin{minipage}[b]{0.242\columnwidth}
		\begin{center}
			\centerline{\includegraphics[width=0.995\columnwidth]{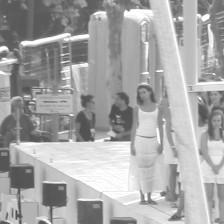}}
			\centerline{\footnotesize{low-crowd(LC)}}
		\end{center}
	\end{minipage}
		\begin{minipage}[b]{0.242\columnwidth}
		\begin{center}
			\centerline{\includegraphics[width=0.995\columnwidth]{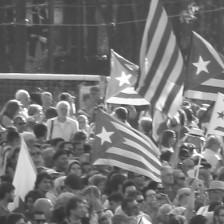}}
			\centerline{\footnotesize{medium-crowd(MC)}}
		\end{center}
	\end{minipage}
	\begin{minipage}[b]{0.242\columnwidth}
		\begin{center}
			\centerline{\includegraphics[width=0.995\columnwidth]{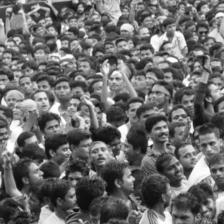}}
			\centerline{\footnotesize{high-crowd(HC)}}
		\end{center}
	\end{minipage}
			
    \vspace{-5mm}
	\caption{\footnotesize{Qualitative results of some test images patches being classified correctly as no-crowd (NC), low-crowd (LC), medium-crowd (MC) or high-crowd (HC) by the CDC classifier as shown for each category column-wise.
	}}
    \vspace{-3mm}
    \label{fig:qualityClassifierResults}
\end{figure}

\begin{table*}[t]\small
	\caption{\footnotesize Computational time analysis of different models.}
	\begin{center}
	\begin{tabular}{|c|c|c|c|c|c|c|}
    \hline
Method & $T_{total}$ (secs) & $T_{avg}$ (secs) & $T_{smallest}$ (secs) & $T_{largest}$ (secs) & MAE & RMSE \\ \hline

CSRNET \cite{li2018csrnet} & \textbf{60.2} & \textbf{0.33} & \textbf{0.09} & \textbf{0.42} & 68.2 & 115.0   \\ \hline
CP-CNN \cite{sindagi2017generating} & 122.8 & 0.68 & 0.31 & 0.85 & 73.6 & 106.4   \\ \hline
ZoomCount-RSE & 85.4 & 0.47 & 0.14 & 0.57 & 66.6 & \textbf{94.5}   \\ \hline
ZoomCount-RFDB & 89.4 & 0.49 & 0.15 & 0.59 & \textbf{66.0} & 97.5  \\ \hline
$ZoomCount-RSE_{pl}$ & 61.9 & 0.34 & - & - & 66.6 & \textbf{94.5} \\ \hline
$ZoomCount-RFDB_{pl}$ & 63.5 & 0.35 & - & - & \textbf{66.0} & 97.5\\ \hline
	\end{tabular}
	\end{center} 
	\label{table:time_analysis}
    \vspace{-1mm}
\end{table*}

\subsection{RFDB Algorithm Selection}
In this paper, we adopt the Random Forest algorithm for the RFDB module. In practice, other machine learning-based classification algorithms can also be employed. In order to choose the best one for our system, we experimented with different classifiers to select the appropriate decision-making algorithm. The results based on the ShanghaiTech and UCF-QNRF datasets are shown in Table \ref{table:ablation_ml}. We observe that ensemble based methods perform better on our relatively smaller and imbalanced RFDB datasets as they prevent over-fitting and high variance by combining several machine learning techniques. After evaluation, the Random Forest appears to be the best choice as the RFDB algorithm as shown in Table \ref{table:ablation_ml}, where the top five best results justify the selection of the Random Forest algorithm. For these experiments, we used machine learning library scikit-learn for python programming \cite{scikitonline}.

\subsection{Qualitative Results}
In Fig. \ref{fig:qualityResults}, we show some good and bad case qualitative results from UCF-QNRF and ShanghaiTech datasets. 
We also compare our results with the ground truth (GT), DenseNet \cite{huang2017densely} Regression (DR) and the state-of-the-art density map methods. In each row, the first three cases demonstrate the good results followed by two bad estimates. The bad case results happen mostly due to the test image being detected as wrong extreme case type by the decision module (DM). We also show some crowd-density classification visual results to demonstrate the qualitative performance of the CDC classifier in Fig. \ref{fig:qualityClassifierResults}.

\subsection{Computational Time Analysis}
In this experiment, we compare the proposed network with two state-of-the-art models (CSRNET \cite{li2018csrnet},CP-CNN \cite{sindagi2017generating}) in terms of real computational time on the ShanghaiTech [14] dataset. The results are shown in Table \ref{table:time_analysis}, where $T_{total}$ and $T_{avg}$ represent the total and average time taken for the whole dataset test images respectively, whereas, $T_{smallest}$ and $T_{largest}$ represent the computation time taken for the smallest and the highest resolution test image respectively in the dataset. For the fairness of comparison, all networks have been evaluated on the same NVIDIA Titan Xp GPU.

As shown in the Table \ref{table:time_analysis}, our method takes a reasonable computation time (in between the two state-of-the-art networks with much closer to the one with the lesser time). However, the proposed method outperforms other models on all standard evaluation metrics. Since our method is modular, we have also shown the pipeline approach ($ZoomCount-RSE_{pl}$, $ZoomCount-RFDB_{pl}$) on the same test images, where next image patches start using the CDC classifier once previous image patches are done using the classifier and now passing through the next modules (the Decision Module and the CRM regressor). We can see that this pipeline approach dramatically decreases the computational time further. This is useful in case of multiple test images. To the best of our knowledge, most of the crowd counting research papers do not provide time analysis, because the main objective for crowd counting in static images is to design a more effective model for counting with lesser error, and less consideration has been given towards the computational time. We believe that efficiency is also a major factor to compare and evaluate different crowd counting models.

\section{Conclusion}
\label{conclusion}
In this work, we have proposed a novel zoom-in and zoom-out based mechanism for effective and accurate crowd counting in highly diverse images. We propose to employ a decision module to detect the extreme high and low dense cases, where most state-of-the-art regression and density map based methods perform worse. The cluttered background regions are also discarded using the rigorous deep CNN 4-way classifier. Even without using any density maps, ZoomCount outperforms the state-of-the-art approaches on four benchmark datasets, thus proving the effectiveness of the proposed model.


%

\ifCLASSOPTIONcaptionsoff
  \newpage
\fi



%





\bibliographystyle{IEEEtranS}  
\bibliography{egbib}  

\end{document}